\documentclass[conference]{IEEEtran}
\IEEEoverridecommandlockouts

\usepackage{cite}
\usepackage{amsmath,amssymb,amsfonts}
\usepackage{graphicx}
\usepackage{booktabs}
\usepackage{array,url}
\usepackage{balance}
\usepackage{multirow}
\usepackage{tabularx}
\usepackage[table]{xcolor}

\begin{document}
\bstctlcite{IEEEtran:BSTcontrol}

\title{D3Tree: Learning to Prune Block-Diffusion Draft Trees for Efficient Verification}
\author{
\IEEEauthorblockN{Hong Nie\textsuperscript{1,\dag}\thanks{\textsuperscript{\dag}This work was achieved during Hong Nie's internship at Unisound AILab under the mentorship of Zhangchi Gao (+1) and Bo Wang (+2).},
Zhangchi Gao\textsuperscript{2},
Jun Yu\textsuperscript{1,*}\thanks{\textsuperscript{*}Corresponding author: harryjun@ustc.edu.cn},
Bo Wang\textsuperscript{2},
Jianqing Sun\textsuperscript{2}
and Shengping Liu\textsuperscript{2}}
\IEEEauthorblockA{\textsuperscript{1}University of Science and Technology of China
\quad\textsuperscript{2}Unisound AI Technology Co., Ltd.\\
gme.niehong@mail.ustc.edu.cn, harryjun@ustc.edu.cn\\
\{gaozhangchi, wangbo, sunjianqing, liushengping\}@unisound.com}
}
\maketitle

% Page numbers in the footer (IEEEtran conference mode omits them by default), as in the Conference paper.
\pagestyle{plain}
\thispagestyle{plain}

\begin{abstract}
Block-diffusion drafters propose a whole block of tokens in one forward pass,
and draft trees built from their position-wise marginals increase the expected
accepted length of speculative decoding. Every tree node, however, occupies a
query row in target verification, so under concurrent serving, low-yield
branches compete with other requests for verification capacity. Existing tree
methods choose the verified nodes by draft probability under a fixed or
cost-selected per-request budget, and thus always retain a prefix of the
drafter's ranking. We present D3Tree, which learns which nodes of a
block-diffusion draft tree are worth verifying. A lightweight tree-size head
with $2H{+}2$ parameters predicts the conditional survival of each candidate
from the draft state, the candidate embedding, and its draft log-probability;
thresholding the product of these scores along parent chains yields an
ancestor-closed subtree whose size adapts to each request and decoding round.
The head is trained on a frozen drafter with conditional labels on the full
candidate tree, and an optional joint mode additionally steers the drafter
toward the retained reference path. At serving time, D3Tree verifies whole
trees under light load and, under heavy load, allocates one node budget across
the batch by predicted survival, sized to the matrix-multiplication tiles of
the target. All retained trees are verified in a single ancestor-masked target
pass, while tree construction, pruning, and scheduling run on the GPU and
overlap with the host. Across six vision-language verifiers and 17 text and
multimodal benchmarks, D3Tree is the fastest of the compared verification
strategies in 70 of 72 load settings, achieving 3.2--4.0$\times$ speedup over
autoregressive decoding for a single request and 1.3--1.8$\times$ at 64
concurrent requests, where fixed-budget trees fall below autoregressive speed.
Under a time-varying arrival trace, it also attains the lowest mean end-to-end
latency in every arrival phase.
\end{abstract}

\begin{IEEEkeywords}
Speculative decoding, block diffusion, draft trees, tree verification,
large language model inference
\end{IEEEkeywords}

\section{Introduction}
\label{sec:introduction}
Autoregressive decoding of large language
models~\cite{vaswani2017attention,brown2020language,touvron2023llama,yang2025qwen3}
is bound by memory bandwidth: every generated token requires a full pass over
the model weights~\cite{pope2023efficiently}. Speculative decoding reduces this
cost by letting an inexpensive drafter propose several tokens that the target
model verifies in a single forward
pass~\cite{leviathan2023speculative,chen2023speculative}. Its speedup is
governed by two quantities: the number of draft tokens accepted per round and
the cost of producing and verifying them. Parallel drafters make this
trade-off explicit, since additional candidates are nearly free to propose,
yet every candidate submitted for verification consumes target-model
capacity.

DFlash~\cite{chen2026dflash} is a block-diffusion drafter that predicts a
block of future tokens in one forward pass, conditioned on target-model
features. Its predictions are position-wise: the distribution at a given
position does not depend on the tokens realized at earlier positions of the
same block. A long linear draft therefore drifts away from the target
continuation, and the drift compounds, because a position contributes only if
all of its predecessors are accepted. As the probability of reaching a deep
position is the product of per-position match probabilities, the accepted
length saturates quickly, however inexpensive the draft.

Draft trees exploit the same parallel output more fully.
DDTree~\cite{ringel2026ddtree} builds the $B$ most probable prefixes from the
position-wise marginals, so that alternative branches can recover where the
top-1 chain fails. The target model, however, must process every node of the
tree although only one path is committed, and each request contributes its
root and all $B$ nodes to the verification batch. Under concurrency,
low-yield branches therefore compete with other requests for verification
capacity, and DDTree itself reports that the accepted length can keep growing
after the speedup has peaked~\cite{ringel2026ddtree}.
CaDDTree~\cite{zhang2026caddtree} selects the budget with a latency model, but
the model describes a single request: the selected tree does not shrink when
concurrent requests compete for verification, and the nodes are still ranked
by the drafter's own probabilities, so the verified set is always a prefix of
that ranking.

\begin{figure*}[t]
 \centering
 \nocite{chen2026dflash,ringel2026ddtree,cheng2026dspark,zhang2026caddtree}
 \includegraphics[width=\linewidth]{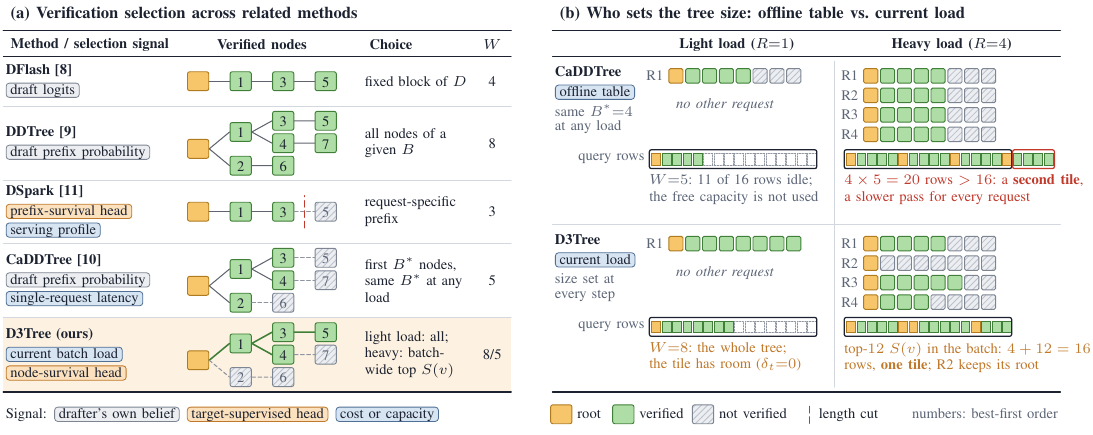}
 \caption{Verification selection in closely related methods. (a)~Each method
 on one candidate space; numbers give the best-first order by draft path
 score, and $W$ counts target query rows, including the root. Chips mark the
 selection signal. (b)~CaDDTree and D3Tree under light load (one request) and
 heavy load ($R=4$ requests in one target pass whose matrix multiplications
 process tiles of $s=16$ rows); below the trees, each step's query rows fill
 the tiles (dashed: idle; red: a second tile). CaDDTree sizes each tree from
 a latency table profiled offline for a single request, regardless of the
 current load, so here it verifies $B^\ast=4$ nodes per request at either
 load: rows stay idle under light load and spill
 into a second tile under heavy load. D3Tree sets the size from the current
 load: the whole tree when the tile has room, and otherwise the 12 nodes of
 highest survival across the batch, which fill one tile.}
 \label{fig:positioning}
\end{figure*}

DSpark~\cite{cheng2026dspark} relies on a different signal. It trains a
confidence head to estimate conditional acceptance and lets a scheduler decide
how long a linear prefix to verify. Learned survival is a natural basis for
deciding how much to verify, but trees raise a question that chains do not:
branches at the same depth can differ in value, so a single length cannot
express which nodes deserve verification. Fig.~\ref{fig:positioning}
contrasts how these methods select the nodes to verify.

We propose \emph{Block \textbf{D}iffusion \textbf{D}ynamic \textbf{D}raft \textbf{Trees}}
(D3Tree), which learns which nodes of a block-diffusion draft tree to verify
and adapts the verification workload to the serving load
(Fig.~\ref{fig:d3tree-overview}). A lightweight \emph{tree-size head}
predicts, for every candidate node, the probability that it matches the
target continuation given that its parent does. Multiplying these scores
along parent chains and thresholding the product yields an ancestor-closed
subtree. Its size equals the number of surviving nodes, so a single
construction budget yields different verification workloads across requests
and decoding rounds, and a confident deep branch can be retained while an
uncertain shallow sibling is pruned.

Whether an additional verified node pays off depends on the load. With few
requests, the target pass is bound by reading the model weights, and the
extra query rows of a whole tree cost little; with many requests, the width of
the batch determines the cost of every pass. D3Tree therefore verifies whole
trees while the batch is small and, beyond a load threshold, retains the
nodes of highest survival across the entire batch, up to a node budget
aligned with the target's matrix-multiplication tiles. All retained trees are
verified in one ancestor-masked target pass, while tree construction,
pruning, and scheduling run on the GPU and overlap with the host.

We evaluate D3Tree on six vision-language verifiers from the Qwen3.5,
InternVL3.5, and Gemma~4 families and on 17 text and multimodal benchmarks,
against linear, dynamically truncated, fixed-budget tree, and cost-selected
tree verification on the same runtime. D3Tree is the fastest strategy in 70 of
72 verifier--benchmark-group--load settings. It reaches
3.2--4.0$\times$ speedup over autoregressive decoding for a single request
and 1.3--1.8$\times$ at 64 concurrent requests, where fixed-budget trees fall
below autoregressive speed (0.26--0.84$\times$), and it is at least
1.15$\times$ as fast as the best cost-selected tree at every load. In serving
experiments, it completes a fixed-concurrency workload first and attains the
lowest mean latency in every phase of a time-varying arrival trace, and it
remains the fastest strategy in most settings on an A800 and an RTX 5090.

Our contributions are as follows.
\begin{itemize}
 \item A conditional-survival tree-size head with $2H{+}2$ parameters that
       extends learned length control from chains to branching trees; a
       single threshold determines both the shape and the size of the
       verified subtree.
 \item A verification-aware training scheme that fits the head on a frozen
       drafter with conditional labels on the full candidate tree, together
       with an optional joint mode that directs token learning toward the
       retained reference path and its first missing position.
 \item A load-dependent serving design that allocates one node budget across
       the batch by learned survival, sized to the target's
       matrix-multiplication tiles, verifies all retained trees in one
       ancestor-masked target pass, and keeps construction, pruning, and
       scheduling on the GPU, overlapped with the host.
 \item An evaluation across six verifiers, 17 benchmarks, concurrency from 1
       to 64, four GPU types, and two serving workloads, including a
       reimplementation of cost-selected trees on the same runtime.
\end{itemize}
Section~\ref{sec:related-work} positions D3Tree against related drafting and
verification schemes, Section~\ref{sec:method} presents the method,
Section~\ref{sec:experiments} evaluates it, and Section~\ref{sec:conclusion}
concludes the paper.

\section{Related Work}
\label{sec:related-work}
\subsection{Speculative Decoding and Drafter Design}
Blockwise parallel decoding~\cite{stern2018blockwise} first predicted several
future tokens and verified them in a single pass of the model. Speculative
decoding and speculative
sampling~\cite{leviathan2023speculative,chen2023speculative,xia2023speculative}
made the scheme lossless by verifying the proposals of a smaller drafter
against the target distribution. Subsequent work improves the drafter through
knowledge distillation~\cite{zhou2024distillspec}, online
adaptation~\cite{liu2024online}, hierarchies of
drafters~\cite{kim2023speculative,spector2023staged,chen2024cascade},
phrase-level drafts~\cite{zhao2024ouroboros}, or retrieval and reference
copying~\cite{he2024rest,yang2023inference}. Draft-free variants accelerate
decoding with Jacobi-style parallel
iterations~\cite{santilli2023parallel,fu2024lookahead,kou2024cllms}, and
SpecTr~\cite{sun2023spectr} verifies several drafts jointly through optimal
transport. Surveys and empirical studies~\cite{xia2024unlocking,yan2025decoding}
show that the end-to-end speedup depends on both the acceptance rate and the
drafting latency, which motivates verifying fewer but better candidates.
A second line of work attaches lightweight draft modules to the target model:
parallel decoding heads in Medusa~\cite{cai2024medusa}, sequentially
dependent heads in Hydra~\cite{ankner2024hydra}, a recurrent
drafter~\cite{cheng2024recurrent}, feature-level autoregression in
EAGLE~\cite{li2024eagle,li2025eagle3}, multi-token prediction
objectives~\cite{gloeckle2024better,deepseek2024v3}, speculative
streaming~\cite{bhendawade2024speculative}, and early-exit
self-speculation~\cite{zhang2024draft,elhoushi2024layerskip,liu2024kangaroo}.
Discrete diffusion language
models~\cite{austin2021structured,sahoo2024simple,nie2025llada} and block
diffusion~\cite{arriola2025block} generate several tokens per step, which
makes them natural parallel drafters~\cite{christopher2025speculative}.
DFlash~\cite{chen2026dflash} predicts a whole block in one pass from target
features; its position-wise distributions make candidate generation
inexpensive but disregard the tokens realized earlier in the block.
DSpark~\cite{cheng2026dspark} adds a lightweight sequential module that models
these within-block transitions, whereas D-PACE~\cite{wu2026dpace} modifies the
training objective, weighting the cross-entropy at each position by a
surrogate of the accepted length. D3Tree retains the parallel proposal and the
D-PACE weighting, and adds supervision tied to the tree that is actually
retained for verification.

\subsection{Draft Trees and Verification Cost}
Tree-based methods distribute the verification budget over alternative
continuations and evaluate them in a single pass under a tree attention
mask~\cite{miao2024specinfer}. Static trees are tuned offline for a given
drafter and hardware~\cite{cai2024medusa,chen2024sequoia}; Sequoia, for
instance, optimizes tree size and depth by dynamic programming under a
hardware-aware cost model. EAGLE-2~\cite{li2024eagle2} and
OPT-Tree~\cite{wang2025opttree} instead adapt the tree to draft confidence at
run time, and SpecExec~\cite{svirschevski2024specexec} constructs very large
trees for offloaded inference. For block-diffusion drafters,
DDTree~\cite{ringel2026ddtree} builds probability-ranked prefixes directly
from the parallel marginals. Its fixed-budget objective measures coverage
under the drafter's factorized distribution, which serves only as a surrogate
for coverage of the target path. CaDDTree~\cite{zhang2026caddtree} trades
this coverage against draft and verification latency to select a budget in
every round, using a latency model profiled for a single request. Both
methods rank nodes by the same drafter probabilities and therefore verify a
prefix of that ranking. D3Tree instead learns a conditional node-survival
signal from reference paths and prunes the bounded tree by accumulated
survival, so its retained nodes need not form a prefix of the draft ranking,
and the number it retains follows the load of the whole batch rather than a
single-request cost. Fig.~\ref{fig:positioning} contrasts these selection
mechanisms.

\begin{figure*}[t]
 \centering
 \includegraphics[width=\linewidth]{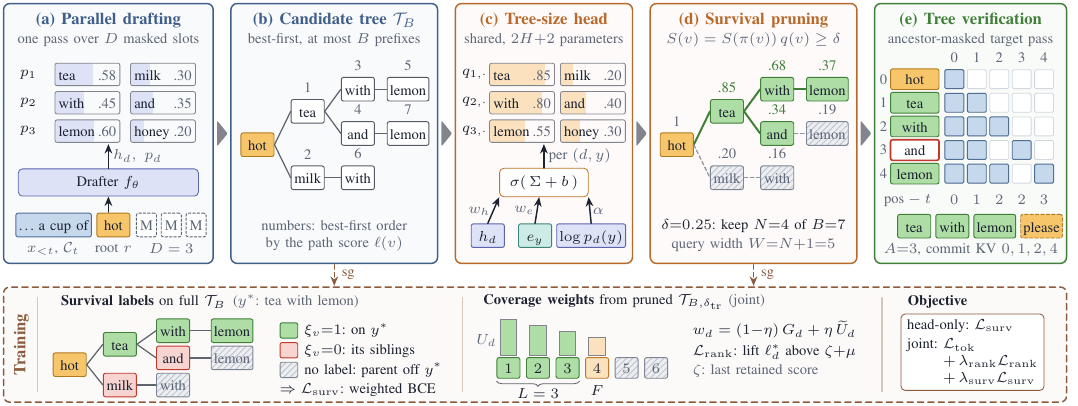}
 \caption{D3Tree in one decoding round (top) and in training (bottom), on a
 running example with $B=7$. (a)~One drafter pass yields position-wise
 marginals $p_d$. (b)~Best-first construction keeps the $B$ highest-scoring
 prefixes; numbers give their order. (c)~The tree-size head scores each
 depth--token pair, so repeated pairs share $q$. (d)~Parent-chain products $S$
 (above the nodes) are thresholded: $N=4$ nodes survive and the target
 verifies $W=5$ rows. (e)~Under the ancestor mask the target accepts
 \emph{tea with lemon}, rejects the retained sibling \emph{and}, emits the
 bonus token, and commits the KV rows of the accepted path. Survival labels
 come from the full tree and are all that head-only fitting uses; in the
 optional joint mode, the coverage $L$ and frontier $F$ of the pruned tree
 weight the token loss. Discrete construction and pruning are stop-gradient
 (sg). The threshold $\delta=0.25$ is illustrative.}
 \label{fig:d3tree-overview}
\end{figure*}

\subsection{Adaptive Verification under Load}
Several methods adapt how much of a linear draft to verify. Dynamic
speculation lookahead~\cite{mamou2024dynamic} trains a classifier that decides
when to stop drafting, SpecDec++~\cite{huang2025specdecpp} adds an
acceptance-prediction head that stops once the predicted rejection risk
exceeds a threshold, and PEARL~\cite{liu2025pearl} adapts the draft length by
overlapping drafting with verification. The confidence scheduler of
DSpark~\cite{cheng2026dspark} is the closest precedent for our approach: it
predicts conditional acceptance, accumulates it along a chain, and lets a
scheduler, rather than a scalar length regressor, determine the verified
prefix. D3Tree extends this representation from chains to branching parent
chains. Its head reads the draft position state, the candidate-token
embedding, and the draft log-probability, and is trained with binary
reference-match labels; the resulting scores determine which branches survive
and hence how many target query rows a request requires. D3Tree retains its
parallel proposal distributions: it neither adopts the sequential transition
module of DSpark nor assumes that DSpark's guarantees for stochastic sampling
carry over to tree pruning.

The amount worth verifying also depends on the load. Serving systems batch
requests at iteration granularity~\cite{yu2022orca}, manage the KV cache in
pages~\cite{kwon2023vllm}, and split long prefills into chunks that share each
step with decoding~\cite{agrawal2024sarathi}. Under such batching, verification
competes with other requests for compute, and the benefit of speculation
diminishes as the load grows~\cite{su2023synergy}. Goodput-oriented
controllers therefore adapt the speculation length to the current
load~\cite{liu2024turbospec}, while long-context serving shifts the trade-off
again because loading the KV cache dominates the step
time~\cite{chen2024magicdec,sun2024triforce}. These systems control the length
of linear drafts. D3Tree controls both the size and the shape of draft trees
from learned node survival, and shares one node budget, aligned with the
target's matrix-multiplication tiles, across the whole batch.

\section{Method}
\label{sec:method}
% Implementation basis and equation-to-source mapping: METHOD_SOURCE_NOTES.md.
% The node-survival path supersedes the earlier depth/bucket formulation.
\subsection{Overview: From Verification Length to Tree Size}
\label{sec:overview}
D3Tree separates three quantities: the draft horizon $D$, the candidate
construction budget $B$, and the retained verification size $N$. The drafter
fixes $D$, the tree builder spends at most $B$ non-root nodes, and the
tree-size head decides how many of them, $N\leq B$, are verified.

For a linear draft of $D$ positions with accepted length $A$,
\begin{equation}
 \mathbb{E}[A]=\sum_{d=1}^{D}\Pr(A\geq d)
             =\sum_{d=1}^{D}\prod_{j=1}^{d}\varphi_j,
 \label{eq:intro-acceptance}
\end{equation}
where $\varphi_j$ is the probability that position $j$ matches given a
matched prefix. Each term is the survival probability of a prefix, which is
what a verification-length rule should estimate. DSpark~\cite{cheng2026dspark}
learns conditional scores $q^{\rm lin}_1,\ldots,q^{\rm lin}_D$ for these
factors; with a threshold, the verified prefix has length
\begin{equation}
 \widehat L_{\delta}=\sum_{d=1}^{D}
 \mathbf{1}\!\left\{\prod_{j=1}^{d}q^{\rm lin}_j\geq\delta\right\}.
 \label{eq:chain-length}
\end{equation}
Equation~\eqref{eq:chain-length} is the chain special case of our pruning
rule, not DSpark's complete serving scheduler, and our tree-size head
generalizes the same conditional-survival representation. Replacing chain prefixes by
tree paths turns the rule into a learned subtree whose cardinality is $N$;
the head thus controls tree size through node survival, without regressing a
depth or a node count. The rest of this section follows the pipeline of
Fig.~\ref{fig:d3tree-overview}: Sections~\ref{sec:candidate-tree}
to~\ref{sec:training} cover drafting, scoring, and training, and
Sections~\ref{sec:admission} to~\ref{sec:overlap} cover the load-dependent
node allocation, tree verification, efficient evaluation of the head, and
the overlapped execution of a decoding round.

\subsection{Parallel Drafting and a Bounded Candidate Tree}
\label{sec:candidate-tree}
D3Tree combines a parallel drafter~\cite{chen2026dflash} with bounded
best-first prefix construction~\cite{ringel2026ddtree}. Let $x_{<t}$ be the
committed context and $r=x_t$ the root token produced by the target model
$P_\phi$, whose vocabulary is $\mathcal{V}_T$. For a horizon $D$, one
forward pass of the drafter $f_\theta$ produces normalized states
$h_d\in\mathbb{R}^{H}$ and position-wise distributions
\begin{align}
 (h_1,\ldots,h_D)
 &=f_\theta(\mathcal{C}_t,r,[\mathrm{MASK}]_{1:D}),
 \label{eq:draft-states}\\
 p_d(a)&=\frac{\exp z_{d,a}}{\sum_{b\in\mathcal{V}_D}\exp z_{d,b}},
 \qquad z_d=\mathrm{LMHead}(h_d).\label{eq:draft-probs}
\end{align}
Here $\mathcal{C}_t$ denotes the cached target features and $\mathcal{V}_D$
the draft vocabulary. A reduced draft vocabulary is mapped to target token
IDs by a fixed injection $\iota$, and we write $p_d(\iota(a))=p_d(a)$.
Because $p_d$ does not depend on the tokens drafted before position $d$, the
proposal factorizes as
\begin{equation}
 Q_{\theta}(y_{1:D}\mid\mathcal{C}_t,r)
   =\prod_{d=1}^{D}p_d(y_d\mid\mathcal{C}_t,r),
 \label{eq:factorized-draft}
\end{equation}
and the same $p_d$ is shared by all branches at depth $d$. The tree uses
these marginals to widen coverage; the learned head later decides which
branches are verified.

Let $\mathcal{K}_d=\operatorname{TopK}(p_d,K)$ with
$K=\min(B,|\mathcal{V}_D|)$, and let $\operatorname{rk}(u)\in\{1,\ldots,K\}$
be the rank of $y_u$ in $\mathcal{K}_{d_u}$. A node $v$ carries a token
$y_v$, a depth $d_v$, a parent $\pi(v)$, and a root-excluding path
$\mathcal{P}(v)$; the root is node~$0$. As in DDTree, a node is scored by
its draft path log-probability,
\begin{equation}
 \ell(v)=\sum_{u\in\mathcal{P}(v)}\log p_{d_u}(y_u),
 \qquad \ell(0)=0,\label{eq:path-score}
\end{equation}
and the bounded tree $\mathcal{T}_B$ consists of the root and the $B$
non-root prefixes with the highest $\ell$. A best-first heap finds them by
lazily pushing the next sibling and the highest-ranked child of each popped
node. Since
\begin{equation}
 \ell(v)=\ell(\pi(v))+\log p_{d_v}(y_v)\leq\ell(\pi(v)),
 \label{eq:tree-order}
\end{equation}
every parent precedes its children; ties
are broken by lexicographic candidate ranks. Top-$K$ extraction keeps the
normalization over $\mathcal{V}_D$ in~\eqref{eq:draft-probs}, and siblings
are not renormalized. The heap runs on the GPU, one program per request, so
neither the candidates nor the trees are copied to the host.

\subsection{A Learned Tree-Size Head via Conditional Survival}
\label{sec:head}
For a candidate $a\in\mathcal{K}_d$ with target-token embedding
$e_a\in\mathbb{R}^{H}$, a shared affine head with $2H{+}2$ parameters
computes
\begin{align}
 g_{d,a}&=w_h^\top h_d+w_e^\top e_a
             +\alpha\log p_d(a)+b,\label{eq:survival-logit}\\
 q_{d,a}&=\sigma(g_{d,a}),\qquad
 q(v)=q_{d_v,y_v}.\label{eq:conditional-survival}
\end{align}
The score $q(v)$ estimates the probability that $v$ matches the reference
continuation given that its parent path matches. Scores are independent
Bernoulli estimates rather than a softmax over siblings. Nodes with the same
depth and token share $q$, but their accumulated scores differ through their
parents.

\begin{figure}[t]
 \centering
 \includegraphics[width=243.99438pt]{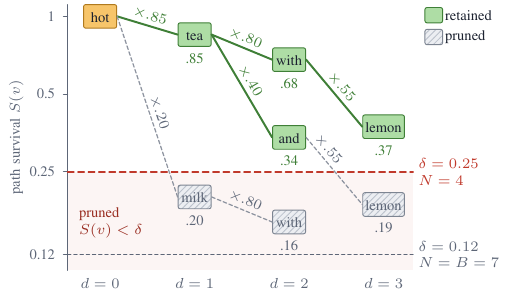}
 \caption{Survival landscape of the example in Fig.~\ref{fig:d3tree-overview}.
 Each node sits at its depth and its path survival $S(v)$ on a log scale;
 edge labels give the conditional score $q$ that multiplies $S$, so no node
 rises above its parent. At $\delta=0.25$, $N=4$ of the $B=7$ candidates
 survive: the depth-3 path is kept, whereas \emph{milk}, second in the
 best-first order, is pruned. The lower threshold $0.12$ keeps a superset,
 here all $B$ nodes. Both thresholds are illustrative;
 Section~\ref{sec:admission} chooses the threshold of each step.}
 \label{fig:d3tree-pruning}
\end{figure}

The root survival and the path survival are defined recursively:
\begin{equation}
 S(0)=1,\qquad S(v)=S(\pi(v))q(v)
       =\prod_{u\in\mathcal{P}(v)}q(u).\label{eq:path-survival}
\end{equation}
For a threshold $\delta$, the retained tree, its size, and its verification
width are
\begin{align}
 \mathcal{T}_{B,\delta}
   &=\{0\}\cup\{v\in\mathcal{T}_B\setminus\{0\}:S(v)\geq\delta\},
   \label{eq:pruning}\\
 N(B,\delta)&=\sum_{v\in\mathcal{T}_B\setminus\{0\}}
        \mathbf{1}\{S(v)\geq\delta\},\label{eq:width}\\
 W(B,\delta)&=N(B,\delta)+1.\label{eq:root-inclusive-width}
\end{align}
Thus $B$ bounds construction, $N$ counts retained draft nodes, and $W$
counts target query rows including the root. Equation~\eqref{eq:width} is
the learned size decision: even at fixed $B$ and $\delta$,
$N=|\mathcal{T}_{B,\delta}|-1$ follows the current node scores. For
$0<\delta\leq1$ the test is additive in log space,
\begin{equation}
 S(v)\geq\delta
 \iff\sum_{u\in\mathcal{P}(v)}-\log q(u)\leq-\log\delta,
 \label{eq:log-survival}
\end{equation}
so a branch survives while its accumulated surprisal stays within the
budget $-\log\delta$. Two invariants follow directly for every non-root
node~$v$:
\begin{align}
 S(v)\leq S(\pi(v)),\quad
 &\bigl[v\in\mathcal{T}_{B,\delta}\Rightarrow
       \pi(v)\in\mathcal{T}_{B,\delta}\bigr],\label{eq:ancestor-closure}\\
 \delta_1\leq\delta_2
 &\Rightarrow\mathcal{T}_{B,\delta_2}\subseteq\mathcal{T}_{B,\delta_1}.
 \label{eq:threshold-nesting}
\end{align}
The first makes every retained tree ancestor-closed; the second makes the
trees for different thresholds nested. Pruning keeps the original node order
and remaps parent indices. Unlike a depth cut, it can keep a deep confident
branch while removing a shallow uncertain one
(Fig.~\ref{fig:d3tree-pruning}); on a single chain it reduces
to~\eqref{eq:chain-length}, that is, $N=\widehat L_\delta$. The number of head
parameters does not depend on $B$. Construction and pruning use different
scores: $\ell$ only bounds which $B$ candidates are built, and $S$ decides
which of them are verified. Section~\ref{sec:admission} chooses the threshold
for every decoding step from the batch.

\subsection{Verification-Aware Training}
\label{sec:training}
\begin{figure*}[t]
 \centering
 \includegraphics[width=\linewidth]{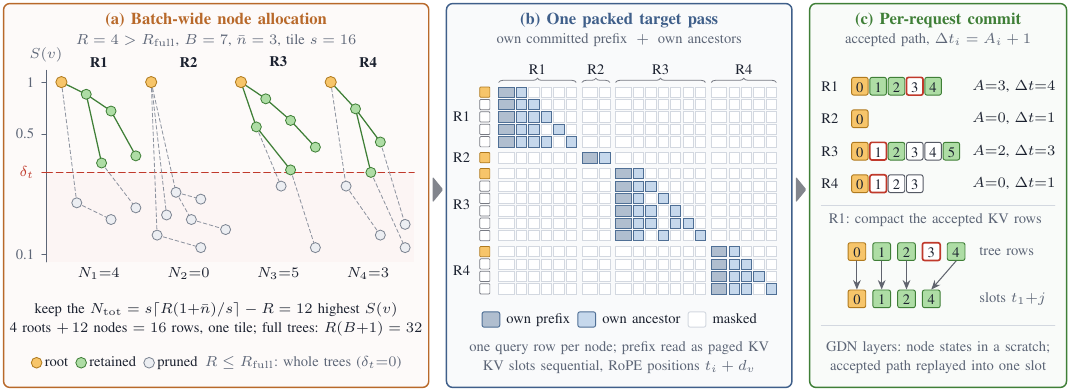}
 \caption{Multi-request serving with D3Tree (illustrative, $R=4$, $B=7$,
 $R_{\rm full}=2$).
 (a)~Every request's full candidate tree on its path survival $S(v)$ (log
 scale; edges join parents and children). Above $R_{\rm full}$ one threshold
 serves the whole batch: $\delta_t$ is the $N_{\rm tot}$-th largest survival,
 with $N_{\rm tot}$ from \eqref{eq:node-budget} (here $\bar n=3$ and a tile of
 $s=16$ rows). Requests keep different numbers of nodes, R2 only its root,
 and the $R+N_{\rm tot}=16$ rows fill one tile instead of the $R(B+1)=32$ of
 full trees. (b)~All retained trees share one target pass. Each query row
 attends to its request's committed prefix through the block table and to
 its own ancestors; rows of other requests are masked. KV slots stay
 sequential while RoPE uses tree depth. (c)~Each request accepts a path
 independently and commits $A_i+1$ tokens; accepted rows are compacted, and
 GDN layers replay the accepted path into the request's state slot.}
 \label{fig:d3tree-serving}
\end{figure*}
The head is fitted on a trained drafter whose weights stay frozen, so it
can be added to an existing block-diffusion drafter without changing its
proposals. All heads in our experiments are trained this way (head-only
fitting). An optional joint mode (Section~\ref{sec:joint}) additionally
adapts the drafter so that useful continuations fall inside the tree that
will be retained. For each training anchor, let $y^*_{1:D}$ be the reference
continuation and $m_d\in\{0,1\}$ its valid-token mask; the root is not
supervised. Head-only fitting takes the prepared response tokens as the
reference. In joint training, the reference tokens are the target head's
greedy predictions at the prepared contexts:
\begin{equation}
 y_d^*=\iota\!\left(\operatorname*{arg\,max}_{a\in\mathcal{V}_D}
 z^{\rm tgt}_{d,a}\right),\qquad
 \bar m_d=\prod_{j=1}^{d}m_j,\label{eq:reference-labels}
\end{equation}
where $z^{\rm tgt}_d$ are the target-head logits, restricted to
$\mathcal{V}_D$ when the draft vocabulary is reduced. In neither case must
every candidate branch be verified online.

\subsubsection{Conditional node supervision}
Let $\mathcal{P}_{y^*}^{B}$ contain the root and the valid reference
prefixes that are present in the \emph{full} candidate tree $\mathcal{T}_B$.
Each candidate node receives a label mask $\chi_v$ and a target $\xi_v$:
\begin{equation}
 \chi_v=\bar m_{d_v}\,
       \mathbf{1}\{\pi(v)\in\mathcal{P}_{y^*}^{B}\},\qquad
 \xi_v=\mathbf{1}\{y_v=y^*_{d_v}\}.\label{eq:conditional-labels}
\end{equation}
A candidate that matches the reference under a matched parent is positive,
its siblings are negative, and descendants of an unmatched parent are not
labeled. With a positive-class weight $w_+>0$ and depth weights
$\omega_d$, the head minimizes the weighted binary cross-entropy
\begin{align}
 \operatorname{BCE}_{w_+}(q,\xi)&=-w_+\xi\log q-(1-\xi)\log(1-q),
       \label{eq:binary-loss}\\
 \mathcal{L}_{\rm surv}
 &=\frac{\sum_{v\ne0}\chi_v\omega_{d_v}\operatorname{BCE}_{w_+}(q(v),\xi_v)}
       {\max\{\sum_{v\ne0}\chi_v\omega_{d_v},\epsilon\}},
       \label{eq:survival-loss}\\
 \omega_d&=\exp(-(d-1)/\gamma),\qquad \gamma>0,
 \label{eq:depth-weight}
\end{align}
where the sums run over the candidates of all anchors and $\omega_d=1$
disables the depth decay. For this loss the log-probability input
of~\eqref{eq:survival-logit} is detached, but gradients still reach $h_d$ in
joint training; target embeddings stay frozen. Because the survival loss
always uses the full tree, a positive that the current head would prune is
still supervised. Head-only fitting minimizes $\mathcal{L}_{\rm surv}$ alone.
Since $q$ does not depend on $B$, a head can be fitted on larger trees than
it serves: ours are fitted on trees with $B=31$, $w_+=1$, and $\gamma=7$,
and serve at $B=15$.

\subsubsection{Optional joint adaptation of the drafter}
\label{sec:joint}
Joint training also updates the drafter, from the first epoch, with a token
loss and a ranking loss that both depend on the tree the current head
retains.

\paragraph{Coverage-aware token weights}
During training, the tree is built and pruned with the current head at a
training threshold $\delta_{\rm tr}$. Let $L$ be the reference coverage of the
retained tree and $F$ the first missing position:
\begin{align}
 L&=\max\bigl\{l\in\{0,\ldots,D\}:\bar m_l=1,
      \ y^*_{1:l}\in\mathcal{T}_{B,\delta_{\rm tr}}\bigr\},
      \label{eq:reference-coverage}\\
 F&=\min(L+1,D),\qquad
 U_d=(D-d+1)\bar m_d\mathbf{1}\{d\leq F\},
 \label{eq:frontier-utility}
\end{align}
where $\bar m_0=1$ and the empty prefix is the root. Tree construction,
pruning, $L$, and all token weights are treated as constants
(stop-gradient). We keep the confidence-based D-PACE
weighting~\cite{wu2026dpace},
\begin{align}
 \kappa_d&=\begin{cases}
 \beta+(1-\beta)\,p_d(y_d^*),&m_d=1,\\
 1,&m_d=0,
 \end{cases}\qquad 0<\beta\leq1,\label{eq:smoothed-confidence}\\
 G_d&=m_d\sum_{j=d}^{D}m_j\prod_{k=1}^{j}\kappa_k,
 \label{eq:dpace-weight}
\end{align}
and blend it with the tree utility:
\begin{equation}
 \widetilde U_d=U_d\frac{\sum_jG_j}{\sum_jU_j},\qquad
 w_d=(1-\eta)G_d+\eta\widetilde U_d,\quad\eta\in[0,1],
 \label{eq:tree-token-weight}
\end{equation}
with $\widetilde U=G$ when $\sum_jU_j=0$. The blend preserves
$\sum_dw_d=\sum_dG_d$ and moves weight toward the covered path and its
frontier. Over anchors $i$, the token loss is
\begin{equation}
 \mathcal{L}_{\rm tok}
 =\frac{\sum_{i,d}m_{i,d}w_{i,d}[-\log p_{i,d}(y^*_{i,d})]}
        {\sum_{i,d}m_{i,d}+\epsilon},
 \label{eq:token-loss}
\end{equation}
whose denominator counts valid tokens rather than their total weight.

\paragraph{Reference-prefix ranking}
The token loss raises the probability of reference tokens, but a reference
prefix enters the tree only if it outranks competing candidates. Let
$\ell^\ast_d$ be the construction score of the reference prefix of length
$d$ and $\zeta$ the detached score of the last retained non-root node in
construction order. With a margin $\mu\geq0$,
\begin{align}
 \ell^\ast_d&=\sum_{j=1}^{d}\log p_j(y^*_j),\qquad
 \zeta=\operatorname{sg}[\ell(v_{\rm last})],\label{eq:ranking-boundary}\\
 \mathcal{L}_{\rm rank}
 &=\frac{1}{|\mathcal{I}|}\sum_{i\in\mathcal{I}}
 \frac{\sum_d U_{i,d}\operatorname{softplus}
       (\zeta_i+\mu-\ell^\ast_{i,d})}
      {\max\{\sum_dU_{i,d},\epsilon\}},\label{eq:ranking-loss}\\
 \mathcal{L}&=\mathcal{L}_{\rm tok}
       +\lambda_{\rm rank}\mathcal{L}_{\rm rank}
       +\lambda_{\rm surv}\mathcal{L}_{\rm surv},
 \label{eq:joint-loss}
\end{align}
where $\mathcal{I}$ contains the anchors with nonzero frontier utility and
an empty set contributes zero. A root-only tree has $\zeta=-\infty$ and thus
zero ranking loss, while its first valid frontier still receives token loss;
$\lambda_{\rm rank}=0$ disables ranking.

The token and ranking losses use the pruned coverage, whereas the
survival loss keeps the full tree. The training threshold $\delta_{\rm tr}$
and the serving thresholds of Section~\ref{sec:admission} are chosen
independently.

\subsection{Load-Dependent Node Allocation}
\label{sec:admission}
Verifying the full trees of $R$ concurrent requests would occupy
\begin{equation}
 W_{\rm full}=R(B+1)
 \label{eq:intro-full-width}
\end{equation}
query rows, one root per request included, whereas the retained trees
occupy $\sum_i(1+N_i)$. Whether verifying more rows pays off depends on
load. With few requests the target pass is bound by reading the model
weights, so a whole tree costs little more than its root, and every pruned
node only loses acceptance. With many requests the width of the batch sets
the cost of the pass, and low-survival nodes delay every request in it.
D3Tree therefore keeps one construction budget $B$ and chooses the threshold
of each decoding step from the number $R$ of requests that verify drafts in
that step:
\begin{equation}
 \delta_t=\begin{cases}
 0, & R\leq R_{\rm full},\\
 S_{(N_{\rm tot})}, & R>R_{\rm full},
 \end{cases}
 \label{eq:step-threshold}
\end{equation}
where $S_{(k)}$ is the $k$-th largest path survival among the non-root nodes
of all $R$ candidate trees and $N_{\rm tot}$ is a batch-wide node budget.
Every request keeps $\mathcal{T}_{B,\delta_t}$ of~\eqref{eq:pruning} with its
own $N_i(B,\delta_t)$ nodes, so each retained tree stays ancestor-closed by
\eqref{eq:ancestor-closure}; nodes tied with $S_{(N_{\rm tot})}$ are all
kept. At light load $\delta_t=0$ retains every node, and D3Tree verifies the
same trees as a fixed-budget tree. Under load the requests compete for one
budget: a request with a confident deep branch keeps more nodes than one
whose candidates are uncertain, and a request whose best candidate falls
below $\delta_t$ is verified with its root alone ($N_i=0$).

The budget follows the cost profile of the target. Its matrix
multiplications process the query rows in tiles of $s$ rows, so the latency
of a target pass is nearly flat within a tile and jumps when the width
crosses a multiple of $s$ (by 12--30\% on an RTX 6000 Ada with $s=64$).
With a per-request allocation $\bar n(R)$ that decreases with load, the
budget fills the last tile that the roots and $\bar n(R)$ nodes per request
reach:
\begin{equation}
 N_{\rm tot}(R)=s\left\lceil\frac{R\,(1+\bar n(R))}{s}\right\rceil-R.
 \label{eq:node-budget}
\end{equation}
We use $B=15$, $s=64$, $R_{\rm full}=8$, and $\bar n(R)=7$ for
$9\leq R<24$ and $\bar n(R)=5$ for $R\geq24$. Hybrid targets with gated
DeltaNet layers~\cite{yang2025gated}, whose verification also advances
recurrent states (Section~\ref{sec:verification}), start pruning earlier:
$R_{\rm full}=4$, $\bar n(R)=7$ for $5\leq R<24$, and $\bar n(R)=5$ for
$R\geq24$. These values are set on the RTX 6000 Ada; on the A800 we retune
$\bar n(R)$ for $R\geq16$ (Section~\ref{sec:device-results}). The budget counts draft
rows only; prefill chunks scheduled in the same iteration-level
batch~\cite{yu2022orca,agrawal2024sarathi} are added on top of it. Without
the allocation table, every step with $R>R_{\rm full}$ applies one fixed
threshold to all requests ($\delta=0.05$ by default). No node floor is used.
D3Tree always verifies trees; it never falls back to linear verification.

\subsection{Tree Verification and State Commitment}
\label{sec:verification}

% Floats placed ahead of their sections so that, at one figure or table per page, none lands among the references.
\begin{table*}[t]
\caption{\normalfont Accepted length $\tau$ per benchmark at concurrency
$R=1$, greedy decoding. All strategies for a verifier share the same drafter and
runtime. DFlash verifies the full linear draft; DSpark a prefix chosen by the
drafter's length head; DDTree the full best-first tree of $B=15$ nodes;
CaDDTree a cost-selected prefix of its best-first tree; D3Tree prunes the
$B=15$ tree under its load-dependent node budget. Benchmarks: GSM8K, MATH-500, AIME25, HumanEval, MBPP, LiveCodeBench,
MT-Bench, OmniDocBench, OCRBench-v2, DocVQA, olmOCR, CharXiv, TextVQA,
MMMU-Pro, GQA, COCO Caption, and MM-MT-Bench. Highest per verifier and
benchmark in bold.}
\label{tab:tau-c1}
\centering
% Generated by tools/gen_experiment_tables.py -- do not edit; the caption and float live in sections/experiments.tex.

\footnotesize
\definecolor{famlin}{RGB}{249,245,252}\definecolor{famtrunc}{RGB}{254,251,229}\definecolor{famtree}{RGB}{240,249,242}\definecolor{famcad}{RGB}{236,244,252}\definecolor{famours}{RGB}{254,244,233}
\setlength{\tabcolsep}{1.0pt}
\renewcommand{\arraystretch}{1.1}
\begin{tabular}{lllccccccccccccccccc}
\toprule
 & & & \multicolumn{7}{c}{Text benchmarks} & \multicolumn{10}{c}{Multimodal benchmarks} \\
\cmidrule(lr){4-10} \cmidrule(lr){11-20}
Model & Strategy & Budget & \multicolumn{3}{c}{Math} & \multicolumn{3}{c}{Code} & \multicolumn{1}{c}{Chat} & \multicolumn{4}{c}{Document \& OCR} & \multicolumn{4}{c}{General vision} & \multicolumn{2}{c}{Capt.\ \& chat} \\
\cmidrule(lr){4-6} \cmidrule(lr){7-9} \cmidrule(lr){10-10} \cmidrule(lr){11-14} \cmidrule(lr){15-18} \cmidrule(lr){19-20}
 & & & GSM8K & MATH & AIME & HEval & MBPP & LCB & MT-B & ODoc & OCR2 & DVQA & olmO & CXiv & TVQA & MMMU & GQA & COCO & MMMT \\
\midrule
\multirow{5}{*}{Q3.5-4B} & \cellcolor{famlin}DFlash & \cellcolor{famlin} & \cellcolor{famlin}5.35 & \cellcolor{famlin}6.63 & \cellcolor{famlin}6.56 & \cellcolor{famlin}5.43 & \cellcolor{famlin}5.12 & \cellcolor{famlin}4.68 & \cellcolor{famlin}3.81 & \cellcolor{famlin}3.08 & \cellcolor{famlin}3.77 & \cellcolor{famlin}4.20 & \cellcolor{famlin}4.86 & \cellcolor{famlin}5.30 & \cellcolor{famlin}3.70 & \cellcolor{famlin}4.33 & \cellcolor{famlin}3.69 & \cellcolor{famlin}3.59 & \cellcolor{famlin}4.29 \\
 & \cellcolor{famtrunc}DSpark & \cellcolor{famtrunc}\multirow{-2}{*}{$N{=}15$} & \cellcolor{famtrunc}4.95 & \cellcolor{famtrunc}6.22 & \cellcolor{famtrunc}5.94 & \cellcolor{famtrunc}5.10 & \cellcolor{famtrunc}4.61 & \cellcolor{famtrunc}4.25 & \cellcolor{famtrunc}3.50 & \cellcolor{famtrunc}2.86 & \cellcolor{famtrunc}3.43 & \cellcolor{famtrunc}3.77 & \cellcolor{famtrunc}4.64 & \cellcolor{famtrunc}4.62 & \cellcolor{famtrunc}3.24 & \cellcolor{famtrunc}3.86 & \cellcolor{famtrunc}3.24 & \cellcolor{famtrunc}3.09 & \cellcolor{famtrunc}3.84 \\
 & \cellcolor{famtree}DDTree & \cellcolor{famtree} & \cellcolor{famtree}\textbf{6.09} & \cellcolor{famtree}7.46 & \cellcolor{famtree}7.59 & \cellcolor{famtree}\textbf{6.25} & \cellcolor{famtree}5.78 & \cellcolor{famtree}5.51 & \cellcolor{famtree}\textbf{4.58} & \cellcolor{famtree}\textbf{3.45} & \cellcolor{famtree}4.43 & \cellcolor{famtree}4.80 & \cellcolor{famtree}5.61 & \cellcolor{famtree}6.00 & \cellcolor{famtree}4.34 & \cellcolor{famtree}\textbf{5.17} & \cellcolor{famtree}4.42 & \cellcolor{famtree}\textbf{4.31} & \cellcolor{famtree}5.07 \\
 & \cellcolor{famcad}CaDDTree & \cellcolor{famcad} & \cellcolor{famcad}6.08 & \cellcolor{famcad}7.42 & \cellcolor{famcad}\textbf{7.70} & \cellcolor{famcad}6.23 & \cellcolor{famcad}5.81 & \cellcolor{famcad}5.49 & \cellcolor{famcad}4.53 & \cellcolor{famcad}3.42 & \cellcolor{famcad}4.44 & \cellcolor{famcad}\textbf{4.82} & \cellcolor{famcad}\textbf{5.64} & \cellcolor{famcad}\textbf{6.01} & \cellcolor{famcad}\textbf{4.35} & \cellcolor{famcad}5.14 & \cellcolor{famcad}4.44 & \cellcolor{famcad}4.29 & \cellcolor{famcad}\textbf{5.10} \\
 & \cellcolor{famours}\textbf{D3Tree} & \cellcolor{famours}\multirow{-3}{*}{$B{=}15$} & \cellcolor{famours}\textbf{6.09} & \cellcolor{famours}\textbf{7.49} & \cellcolor{famours}7.39 & \cellcolor{famours}\textbf{6.25} & \cellcolor{famours}\textbf{5.86} & \cellcolor{famours}\textbf{5.53} & \cellcolor{famours}4.52 & \cellcolor{famours}3.44 & \cellcolor{famours}\textbf{4.49} & \cellcolor{famours}4.76 & \cellcolor{famours}5.61 & \cellcolor{famours}5.97 & \cellcolor{famours}4.34 & \cellcolor{famours}5.11 & \cellcolor{famours}\textbf{4.46} & \cellcolor{famours}4.27 & \cellcolor{famours}5.04 \\
\cmidrule(l){1-20}
\multirow{5}{*}{Q3.5-9B} & \cellcolor{famlin}DFlash & \cellcolor{famlin} & \cellcolor{famlin}5.14 & \cellcolor{famlin}6.35 & \cellcolor{famlin}6.36 & \cellcolor{famlin}5.22 & \cellcolor{famlin}4.95 & \cellcolor{famlin}4.71 & \cellcolor{famlin}3.74 & \cellcolor{famlin}2.91 & \cellcolor{famlin}3.76 & \cellcolor{famlin}4.08 & \cellcolor{famlin}4.79 & \cellcolor{famlin}4.99 & \cellcolor{famlin}3.51 & \cellcolor{famlin}4.16 & \cellcolor{famlin}3.61 & \cellcolor{famlin}3.45 & \cellcolor{famlin}4.19 \\
 & \cellcolor{famtrunc}DSpark & \cellcolor{famtrunc}\multirow{-2}{*}{$N{=}15$} & \cellcolor{famtrunc}4.64 & \cellcolor{famtrunc}5.68 & \cellcolor{famtrunc}5.61 & \cellcolor{famtrunc}4.80 & \cellcolor{famtrunc}4.53 & \cellcolor{famtrunc}4.17 & \cellcolor{famtrunc}3.34 & \cellcolor{famtrunc}2.69 & \cellcolor{famtrunc}3.37 & \cellcolor{famtrunc}3.71 & \cellcolor{famtrunc}4.34 & \cellcolor{famtrunc}4.50 & \cellcolor{famtrunc}3.02 & \cellcolor{famtrunc}3.70 & \cellcolor{famtrunc}3.09 & \cellcolor{famtrunc}3.01 & \cellcolor{famtrunc}3.72 \\
 & \cellcolor{famtree}DDTree & \cellcolor{famtree} & \cellcolor{famtree}5.93 & \cellcolor{famtree}\textbf{7.17} & \cellcolor{famtree}\textbf{7.19} & \cellcolor{famtree}6.10 & \cellcolor{famtree}5.81 & \cellcolor{famtree}\textbf{5.66} & \cellcolor{famtree}\textbf{4.51} & \cellcolor{famtree}3.20 & \cellcolor{famtree}\textbf{4.47} & \cellcolor{famtree}\textbf{4.75} & \cellcolor{famtree}\textbf{5.51} & \cellcolor{famtree}\textbf{5.82} & \cellcolor{famtree}\textbf{4.21} & \cellcolor{famtree}\textbf{5.00} & \cellcolor{famtree}4.30 & \cellcolor{famtree}\textbf{4.22} & \cellcolor{famtree}\textbf{4.98} \\
 & \cellcolor{famcad}CaDDTree & \cellcolor{famcad} & \cellcolor{famcad}5.97 & \cellcolor{famcad}7.10 & \cellcolor{famcad}7.14 & \cellcolor{famcad}6.06 & \cellcolor{famcad}\textbf{5.91} & \cellcolor{famcad}5.54 & \cellcolor{famcad}4.48 & \cellcolor{famcad}\textbf{3.25} & \cellcolor{famcad}4.43 & \cellcolor{famcad}4.70 & \cellcolor{famcad}5.45 & \cellcolor{famcad}5.75 & \cellcolor{famcad}4.08 & \cellcolor{famcad}4.99 & \cellcolor{famcad}4.33 & \cellcolor{famcad}4.19 & \cellcolor{famcad}4.92 \\
 & \cellcolor{famours}\textbf{D3Tree} & \cellcolor{famours}\multirow{-3}{*}{$B{=}15$} & \cellcolor{famours}\textbf{6.04} & \cellcolor{famours}7.09 & \cellcolor{famours}7.16 & \cellcolor{famours}\textbf{6.11} & \cellcolor{famours}5.82 & \cellcolor{famours}5.59 & \cellcolor{famours}4.49 & \cellcolor{famours}3.23 & \cellcolor{famours}4.41 & \cellcolor{famours}4.74 & \cellcolor{famours}5.45 & \cellcolor{famours}5.75 & \cellcolor{famours}4.15 & \cellcolor{famours}4.92 & \cellcolor{famours}\textbf{4.35} & \cellcolor{famours}4.21 & \cellcolor{famours}4.95 \\
\cmidrule(l){1-20}
\multirow{5}{*}{IVL3.5-4B} & \cellcolor{famlin}DFlash & \cellcolor{famlin} & \cellcolor{famlin}5.49 & \cellcolor{famlin}7.27 & \cellcolor{famlin}4.80 & \cellcolor{famlin}3.82 & \cellcolor{famlin}4.20 & \cellcolor{famlin}3.76 & \cellcolor{famlin}3.32 & \cellcolor{famlin}2.98 & \cellcolor{famlin}3.14 & \cellcolor{famlin}3.59 & \cellcolor{famlin}4.68 & \cellcolor{famlin}4.67 & \cellcolor{famlin}2.64 & \cellcolor{famlin}5.23 & \cellcolor{famlin}3.35 & \cellcolor{famlin}3.02 & \cellcolor{famlin}3.60 \\
 & \cellcolor{famtrunc}DSpark & \cellcolor{famtrunc}\multirow{-2}{*}{$N{=}15$} & \cellcolor{famtrunc}5.47 & \cellcolor{famtrunc}7.00 & \cellcolor{famtrunc}4.15 & \cellcolor{famtrunc}3.46 & \cellcolor{famtrunc}3.89 & \cellcolor{famtrunc}3.36 & \cellcolor{famtrunc}2.97 & \cellcolor{famtrunc}2.96 & \cellcolor{famtrunc}3.07 & \cellcolor{famtrunc}2.98 & \cellcolor{famtrunc}4.21 & \cellcolor{famtrunc}4.08 & \cellcolor{famtrunc}2.28 & \cellcolor{famtrunc}4.95 & \cellcolor{famtrunc}2.74 & \cellcolor{famtrunc}2.52 & \cellcolor{famtrunc}3.13 \\
 & \cellcolor{famtree}DDTree & \cellcolor{famtree} & \cellcolor{famtree}6.39 & \cellcolor{famtree}\textbf{8.05} & \cellcolor{famtree}5.51 & \cellcolor{famtree}4.57 & \cellcolor{famtree}\textbf{5.04} & \cellcolor{famtree}\textbf{4.61} & \cellcolor{famtree}4.01 & \cellcolor{famtree}3.48 & \cellcolor{famtree}\textbf{3.47} & \cellcolor{famtree}4.19 & \cellcolor{famtree}\textbf{5.48} & \cellcolor{famtree}\textbf{5.57} & \cellcolor{famtree}\textbf{3.30} & \cellcolor{famtree}6.27 & \cellcolor{famtree}4.04 & \cellcolor{famtree}3.77 & \cellcolor{famtree}4.24 \\
 & \cellcolor{famcad}CaDDTree & \cellcolor{famcad} & \cellcolor{famcad}6.34 & \cellcolor{famcad}8.04 & \cellcolor{famcad}5.50 & \cellcolor{famcad}4.56 & \cellcolor{famcad}5.01 & \cellcolor{famcad}4.57 & \cellcolor{famcad}4.03 & \cellcolor{famcad}\textbf{3.58} & \cellcolor{famcad}3.46 & \cellcolor{famcad}\textbf{4.22} & \cellcolor{famcad}5.29 & \cellcolor{famcad}5.39 & \cellcolor{famcad}3.21 & \cellcolor{famcad}\textbf{6.28} & \cellcolor{famcad}4.07 & \cellcolor{famcad}3.75 & \cellcolor{famcad}\textbf{4.34} \\
 & \cellcolor{famours}\textbf{D3Tree} & \cellcolor{famours}\multirow{-3}{*}{$B{=}15$} & \cellcolor{famours}\textbf{6.48} & \cellcolor{famours}8.04 & \cellcolor{famours}\textbf{5.52} & \cellcolor{famours}\textbf{4.59} & \cellcolor{famours}5.01 & \cellcolor{famours}4.56 & \cellcolor{famours}\textbf{4.06} & \cellcolor{famours}3.39 & \cellcolor{famours}3.42 & \cellcolor{famours}4.17 & \cellcolor{famours}5.40 & \cellcolor{famours}5.31 & \cellcolor{famours}\textbf{3.30} & \cellcolor{famours}6.15 & \cellcolor{famours}\textbf{4.08} & \cellcolor{famours}\textbf{3.78} & \cellcolor{famours}4.26 \\
\cmidrule(l){1-20}
\multirow{5}{*}{IVL3.5-8B} & \cellcolor{famlin}DFlash & \cellcolor{famlin} & \cellcolor{famlin}7.08 & \cellcolor{famlin}6.14 & \cellcolor{famlin}5.43 & \cellcolor{famlin}5.38 & \cellcolor{famlin}5.04 & \cellcolor{famlin}5.29 & \cellcolor{famlin}3.16 & \cellcolor{famlin}3.21 & \cellcolor{famlin}3.46 & \cellcolor{famlin}3.69 & \cellcolor{famlin}4.26 & \cellcolor{famlin}4.75 & \cellcolor{famlin}2.68 & \cellcolor{famlin}4.75 & \cellcolor{famlin}3.31 & \cellcolor{famlin}2.93 & \cellcolor{famlin}3.73 \\
 & \cellcolor{famtrunc}DSpark & \cellcolor{famtrunc}\multirow{-2}{*}{$N{=}15$} & \cellcolor{famtrunc}6.80 & \cellcolor{famtrunc}6.03 & \cellcolor{famtrunc}5.04 & \cellcolor{famtrunc}5.13 & \cellcolor{famtrunc}4.74 & \cellcolor{famtrunc}5.09 & \cellcolor{famtrunc}2.93 & \cellcolor{famtrunc}3.09 & \cellcolor{famtrunc}3.52 & \cellcolor{famtrunc}3.15 & \cellcolor{famtrunc}3.88 & \cellcolor{famtrunc}4.40 & \cellcolor{famtrunc}2.34 & \cellcolor{famtrunc}4.35 & \cellcolor{famtrunc}2.78 & \cellcolor{famtrunc}2.49 & \cellcolor{famtrunc}3.41 \\
 & \cellcolor{famtree}DDTree & \cellcolor{famtree} & \cellcolor{famtree}7.64 & \cellcolor{famtree}\textbf{7.01} & \cellcolor{famtree}6.15 & \cellcolor{famtree}6.24 & \cellcolor{famtree}5.76 & \cellcolor{famtree}6.08 & \cellcolor{famtree}3.80 & \cellcolor{famtree}3.30 & \cellcolor{famtree}\textbf{4.08} & \cellcolor{famtree}4.23 & \cellcolor{famtree}4.88 & \cellcolor{famtree}\textbf{5.55} & \cellcolor{famtree}3.35 & \cellcolor{famtree}5.49 & \cellcolor{famtree}\textbf{4.11} & \cellcolor{famtree}3.67 & \cellcolor{famtree}4.47 \\
 & \cellcolor{famcad}CaDDTree & \cellcolor{famcad} & \cellcolor{famcad}7.66 & \cellcolor{famcad}6.82 & \cellcolor{famcad}6.02 & \cellcolor{famcad}6.28 & \cellcolor{famcad}\textbf{5.89} & \cellcolor{famcad}\textbf{6.12} & \cellcolor{famcad}\textbf{3.92} & \cellcolor{famcad}3.21 & \cellcolor{famcad}3.83 & \cellcolor{famcad}\textbf{4.38} & \cellcolor{famcad}4.92 & \cellcolor{famcad}5.53 & \cellcolor{famcad}3.35 & \cellcolor{famcad}5.46 & \cellcolor{famcad}4.05 & \cellcolor{famcad}\textbf{3.69} & \cellcolor{famcad}\textbf{4.48} \\
 & \cellcolor{famours}\textbf{D3Tree} & \cellcolor{famours}\multirow{-3}{*}{$B{=}15$} & \cellcolor{famours}\textbf{7.68} & \cellcolor{famours}7.00 & \cellcolor{famours}\textbf{6.21} & \cellcolor{famours}\textbf{6.35} & \cellcolor{famours}5.88 & \cellcolor{famours}6.10 & \cellcolor{famours}3.86 & \cellcolor{famours}\textbf{3.63} & \cellcolor{famours}3.85 & \cellcolor{famours}4.25 & \cellcolor{famours}\textbf{5.02} & \cellcolor{famours}5.49 & \cellcolor{famours}\textbf{3.38} & \cellcolor{famours}\textbf{5.52} & \cellcolor{famours}4.06 & \cellcolor{famours}3.68 & \cellcolor{famours}4.47 \\
\cmidrule(l){1-20}
\multirow{5}{*}{G4-E2B} & \cellcolor{famlin}DFlash & \cellcolor{famlin} & \cellcolor{famlin}5.76 & \cellcolor{famlin}6.68 & \cellcolor{famlin}6.36 & \cellcolor{famlin}5.71 & \cellcolor{famlin}5.19 & \cellcolor{famlin}4.73 & \cellcolor{famlin}2.94 & \cellcolor{famlin}2.35 & \cellcolor{famlin}3.56 & \cellcolor{famlin}3.62 & \cellcolor{famlin}3.08 & \cellcolor{famlin}4.30 & \cellcolor{famlin}2.75 & \cellcolor{famlin}3.77 & \cellcolor{famlin}3.04 & \cellcolor{famlin}2.93 & \cellcolor{famlin}3.09 \\
 & \cellcolor{famtrunc}DSpark & \cellcolor{famtrunc}\multirow{-2}{*}{$N{=}15$} & \cellcolor{famtrunc}5.20 & \cellcolor{famtrunc}6.23 & \cellcolor{famtrunc}5.68 & \cellcolor{famtrunc}5.36 & \cellcolor{famtrunc}4.74 & \cellcolor{famtrunc}4.27 & \cellcolor{famtrunc}2.72 & \cellcolor{famtrunc}2.37 & \cellcolor{famtrunc}3.30 & \cellcolor{famtrunc}3.14 & \cellcolor{famtrunc}2.97 & \cellcolor{famtrunc}3.83 & \cellcolor{famtrunc}2.46 & \cellcolor{famtrunc}3.25 & \cellcolor{famtrunc}2.57 & \cellcolor{famtrunc}2.51 & \cellcolor{famtrunc}2.71 \\
 & \cellcolor{famtree}DDTree & \cellcolor{famtree} & \cellcolor{famtree}\textbf{6.55} & \cellcolor{famtree}7.45 & \cellcolor{famtree}\textbf{7.34} & \cellcolor{famtree}\textbf{6.49} & \cellcolor{famtree}5.93 & \cellcolor{famtree}\textbf{5.59} & \cellcolor{famtree}3.64 & \cellcolor{famtree}2.72 & \cellcolor{famtree}3.95 & \cellcolor{famtree}\textbf{4.22} & \cellcolor{famtree}3.65 & \cellcolor{famtree}5.05 & \cellcolor{famtree}3.44 & \cellcolor{famtree}\textbf{4.55} & \cellcolor{famtree}\textbf{3.84} & \cellcolor{famtree}\textbf{3.71} & \cellcolor{famtree}3.75 \\
 & \cellcolor{famcad}CaDDTree & \cellcolor{famcad} & \cellcolor{famcad}6.54 & \cellcolor{famcad}7.50 & \cellcolor{famcad}6.97 & \cellcolor{famcad}6.43 & \cellcolor{famcad}\textbf{5.94} & \cellcolor{famcad}5.53 & \cellcolor{famcad}3.66 & \cellcolor{famcad}2.70 & \cellcolor{famcad}\textbf{3.99} & \cellcolor{famcad}4.19 & \cellcolor{famcad}\textbf{3.73} & \cellcolor{famcad}5.09 & \cellcolor{famcad}3.47 & \cellcolor{famcad}4.53 & \cellcolor{famcad}3.82 & \cellcolor{famcad}3.70 & \cellcolor{famcad}\textbf{3.77} \\
 & \cellcolor{famours}\textbf{D3Tree} & \cellcolor{famours}\multirow{-3}{*}{$B{=}15$} & \cellcolor{famours}6.53 & \cellcolor{famours}\textbf{7.53} & \cellcolor{famours}7.03 & \cellcolor{famours}6.48 & \cellcolor{famours}5.90 & \cellcolor{famours}5.57 & \cellcolor{famours}\textbf{3.68} & \cellcolor{famours}\textbf{2.84} & \cellcolor{famours}3.84 & \cellcolor{famours}4.21 & \cellcolor{famours}3.68 & \cellcolor{famours}\textbf{5.10} & \cellcolor{famours}\textbf{3.48} & \cellcolor{famours}4.54 & \cellcolor{famours}3.80 & \cellcolor{famours}\textbf{3.71} & \cellcolor{famours}\textbf{3.77} \\
\cmidrule(l){1-20}
\multirow{5}{*}{G4-E4B} & \cellcolor{famlin}DFlash & \cellcolor{famlin} & \cellcolor{famlin}6.32 & \cellcolor{famlin}7.13 & \cellcolor{famlin}6.99 & \cellcolor{famlin}6.49 & \cellcolor{famlin}5.64 & \cellcolor{famlin}5.16 & \cellcolor{famlin}3.13 & \cellcolor{famlin}2.51 & \cellcolor{famlin}3.49 & \cellcolor{famlin}3.71 & \cellcolor{famlin}3.43 & \cellcolor{famlin}4.62 & \cellcolor{famlin}2.97 & \cellcolor{famlin}4.29 & \cellcolor{famlin}3.23 & \cellcolor{famlin}2.98 & \cellcolor{famlin}3.31 \\
 & \cellcolor{famtrunc}DSpark & \cellcolor{famtrunc}\multirow{-2}{*}{$N{=}15$} & \cellcolor{famtrunc}5.77 & \cellcolor{famtrunc}6.71 & \cellcolor{famtrunc}6.50 & \cellcolor{famtrunc}6.26 & \cellcolor{famtrunc}5.22 & \cellcolor{famtrunc}4.84 & \cellcolor{famtrunc}2.87 & \cellcolor{famtrunc}2.52 & \cellcolor{famtrunc}3.21 & \cellcolor{famtrunc}2.84 & \cellcolor{famtrunc}3.17 & \cellcolor{famtrunc}4.06 & \cellcolor{famtrunc}2.37 & \cellcolor{famtrunc}3.76 & \cellcolor{famtrunc}2.60 & \cellcolor{famtrunc}2.50 & \cellcolor{famtrunc}2.85 \\
 & \cellcolor{famtree}DDTree & \cellcolor{famtree} & \cellcolor{famtree}\textbf{7.03} & \cellcolor{famtree}\textbf{7.90} & \cellcolor{famtree}\textbf{7.97} & \cellcolor{famtree}7.02 & \cellcolor{famtree}\textbf{6.46} & \cellcolor{famtree}5.93 & \cellcolor{famtree}\textbf{3.85} & \cellcolor{famtree}2.86 & \cellcolor{famtree}3.96 & \cellcolor{famtree}4.34 & \cellcolor{famtree}\textbf{4.06} & \cellcolor{famtree}\textbf{5.43} & \cellcolor{famtree}\textbf{3.59} & \cellcolor{famtree}4.99 & \cellcolor{famtree}\textbf{4.10} & \cellcolor{famtree}\textbf{3.75} & \cellcolor{famtree}\textbf{4.03} \\
 & \cellcolor{famcad}CaDDTree & \cellcolor{famcad} & \cellcolor{famcad}6.97 & \cellcolor{famcad}7.86 & \cellcolor{famcad}7.93 & \cellcolor{famcad}7.05 & \cellcolor{famcad}6.40 & \cellcolor{famcad}5.93 & \cellcolor{famcad}3.83 & \cellcolor{famcad}\textbf{2.89} & \cellcolor{famcad}4.00 & \cellcolor{famcad}4.24 & \cellcolor{famcad}\textbf{4.06} & \cellcolor{famcad}5.42 & \cellcolor{famcad}\textbf{3.59} & \cellcolor{famcad}5.07 & \cellcolor{famcad}4.02 & \cellcolor{famcad}3.74 & \cellcolor{famcad}3.97 \\
 & \cellcolor{famours}\textbf{D3Tree} & \cellcolor{famours}\multirow{-3}{*}{$B{=}15$} & \cellcolor{famours}7.00 & \cellcolor{famours}\textbf{7.90} & \cellcolor{famours}7.76 & \cellcolor{famours}\textbf{7.08} & \cellcolor{famours}6.43 & \cellcolor{famours}\textbf{5.96} & \cellcolor{famours}3.84 & \cellcolor{famours}2.87 & \cellcolor{famours}\textbf{4.02} & \cellcolor{famours}\textbf{4.35} & \cellcolor{famours}4.03 & \cellcolor{famours}5.40 & \cellcolor{famours}\textbf{3.59} & \cellcolor{famours}\textbf{5.08} & \cellcolor{famours}4.06 & \cellcolor{famours}3.74 & \cellcolor{famours}4.00 \\
\bottomrule
\end{tabular}

\end{table*}

\begin{table*}[t]
\caption{\normalfont Speedup over AR per benchmark at concurrency $R=1$:
output throughput divided by that of AR for the same verifier, GPU, and
prompts. Strategies and benchmark abbreviations as in
Table~\ref{tab:tau-c1}. Best per verifier and benchmark
in bold.}
\label{tab:speedup-c1}
\centering
% Generated by tools/gen_experiment_tables.py -- do not edit; the caption and float live in sections/experiments.tex.

\footnotesize
\definecolor{famlin}{RGB}{249,245,252}\definecolor{famtrunc}{RGB}{254,251,229}\definecolor{famtree}{RGB}{240,249,242}\definecolor{famcad}{RGB}{236,244,252}\definecolor{famours}{RGB}{254,244,233}
\setlength{\tabcolsep}{1.0pt}
\renewcommand{\arraystretch}{1.1}
\begin{tabular}{lllccccccccccccccccc}
\toprule
 & & & \multicolumn{7}{c}{Text benchmarks} & \multicolumn{10}{c}{Multimodal benchmarks} \\
\cmidrule(lr){4-10} \cmidrule(lr){11-20}
Model & Strategy & Budget & \multicolumn{3}{c}{Math} & \multicolumn{3}{c}{Code} & \multicolumn{1}{c}{Chat} & \multicolumn{4}{c}{Document \& OCR} & \multicolumn{4}{c}{General vision} & \multicolumn{2}{c}{Capt.\ \& chat} \\
\cmidrule(lr){4-6} \cmidrule(lr){7-9} \cmidrule(lr){10-10} \cmidrule(lr){11-14} \cmidrule(lr){15-18} \cmidrule(lr){19-20}
 & & & GSM8K & MATH & AIME & HEval & MBPP & LCB & MT-B & ODoc & OCR2 & DVQA & olmO & CXiv & TVQA & MMMU & GQA & COCO & MMMT \\
\midrule
\multirow{5}{*}{Q3.5-4B} & \cellcolor{famlin}DFlash & \cellcolor{famlin} & \cellcolor{famlin}3.38$\times$ & \cellcolor{famlin}4.07$\times$ & \cellcolor{famlin}4.07$\times$ & \cellcolor{famlin}3.40$\times$ & \cellcolor{famlin}3.23$\times$ & \cellcolor{famlin}2.92$\times$ & \cellcolor{famlin}2.40$\times$ & \cellcolor{famlin}1.99$\times$ & \cellcolor{famlin}2.41$\times$ & \cellcolor{famlin}2.40$\times$ & \cellcolor{famlin}3.08$\times$ & \cellcolor{famlin}3.26$\times$ & \cellcolor{famlin}2.32$\times$ & \cellcolor{famlin}2.71$\times$ & \cellcolor{famlin}2.32$\times$ & \cellcolor{famlin}2.25$\times$ & \cellcolor{famlin}2.64$\times$ \\
 & \cellcolor{famtrunc}DSpark & \cellcolor{famtrunc}\multirow{-2}{*}{$N{=}15$} & \cellcolor{famtrunc}3.25$\times$ & \cellcolor{famtrunc}3.98$\times$ & \cellcolor{famtrunc}3.79$\times$ & \cellcolor{famtrunc}3.37$\times$ & \cellcolor{famtrunc}2.97$\times$ & \cellcolor{famtrunc}2.74$\times$ & \cellcolor{famtrunc}2.24$\times$ & \cellcolor{famtrunc}1.88$\times$ & \cellcolor{famtrunc}2.24$\times$ & \cellcolor{famtrunc}2.27$\times$ & \cellcolor{famtrunc}3.00$\times$ & \cellcolor{famtrunc}2.95$\times$ & \cellcolor{famtrunc}2.07$\times$ & \cellcolor{famtrunc}2.49$\times$ & \cellcolor{famtrunc}2.12$\times$ & \cellcolor{famtrunc}2.03$\times$ & \cellcolor{famtrunc}2.48$\times$ \\
\bottomrule
\end{tabular}

\end{table*}

Let $\widetilde{\mathcal{T}}$ be an admitted tree and $\mathcal{A}^{+}(v)$
the root-inclusive ancestors of $v$, including $v$ itself. As in tree-based
speculative inference~\cite{miao2024specinfer}, the target assigns
depth-based rotary positions~\cite{su2024roformer} and applies the tree
attention mask
\begin{equation}
 \operatorname{pos}(v)=t+d_v,\quad d_0=0,\qquad
 M_{vu}=\begin{cases}
 0,&u\in\mathcal{A}^{+}(v),\\
 -\infty,&\text{otherwise}.
 \end{cases}\label{eq:tree-mask}
\end{equation}
In addition, every node attends to $x_{<t}$ under the target's native
attention pattern. In a batch, the trees of all requests are packed into one
variable-length query. Each query row sees only the committed prefix of its
own request, read from the paged KV cache~\cite{kwon2023vllm}, and its own
ancestors (Fig.~\ref{fig:d3tree-serving}(b)); KV entries are written to
consecutive slots, whereas rotary positions follow~\eqref{eq:tree-mask}.
Tree rows run a Triton kernel derived from vLLM's unified paged attention:
$M$ enters as an additive bias over each request's window of at most
$B{+}1$ rows, and for up to 256 tree rows the prefix is additionally split
into segments along the KV length. No
per-step block mask is built, and the kernel reads only persistent buffers,
so the whole target pass can be captured as a CUDA graph. When prefill
chunks share the step, the tree rows, which the batch places first, take
this kernel in one launch and the chunks a causal one. Only queries inside a
bidirectional image span of a multimodal target fall back to a block-sparse
FlexAttention kernel~\cite{dong2024flex}. We consider greedy verification;
path-dependent penalties, masks, and logits processors are not supported.
The target prediction after node $v$ is
\begin{equation}
 \hat y(v)=\operatorname*{arg\,max}_{a\in\mathcal{V}_T}
 P_\phi\bigl(a\mid x_{<t},r,y_{\mathcal{P}(v)}\bigr).
 \label{eq:target-verification}
\end{equation}
Starting from $v_0=0$, verification follows a child only if its token equals
the current target prediction:
\begin{equation}
 \begin{aligned}
 v_{j+1}&=\operatorname{child}(v_j,\hat y(v_j))&&\text{if present},\\
 b_{\rm next}&=\hat y(v_A)&&\text{at the first miss}.
 \end{aligned}
 \label{eq:accept-path}
\end{equation}
The round emits the $A$ accepted draft tokens followed by $b_{\rm next}$,
which becomes the next root. The KV entries of the accepted path are
compacted into the committed cache:
\begin{equation}
 \begin{aligned}
 \mathrm{KV}_{\rm new}[t+j]&=\mathrm{KV}_{\rm tree}[v_j],\quad j=0,\ldots,A,\\
 \Delta t&=A+1.
 \end{aligned}
 \label{eq:kv-commit}
\end{equation}
Because the accepted path need not be contiguous in the packed tree,
compaction follows node indices; it moves the accepted rows of all layers
that share a KV cache group in one gather and one scatter. Recurrent layers
of hybrid targets, such as gated DeltaNet~\cite{yang2025gated} or other
state-space layers~\cite{gu2023mamba}, keep a single state slot per request.
The target pass reads it without writing it; rows whose retained tree
branches keep the state of every non-root node in a shared scratch, each
computed from its parent's state, whereas chain-shaped rows need none. Once
the accepted path is known, its tokens are replayed from the slot with the
same arithmetic, order, and tiling as the target pass, so the committed
state is bitwise the state of the accepted branch, and the newest
convolution inputs of the path are written to the convolution window.

If tree evaluation reproduces the sequential target logits and state
transitions, \eqref{eq:accept-path} yields the same greedy continuation as
autoregressive decoding, by induction over the accepted path. This is a
conditional property: batched finite-precision kernels require separate
validation of numerical parity. The survival head only selects work; every
emitted token is determined by the target's predictions.

\subsection{Cached and Fused Survival Evaluation}
\label{sec:fused}

% Floats placed ahead of their sections so that, at one figure or table per page, none lands among the references.
\begin{table*}[t]
\caption{\normalfont Speedup over AR at concurrency $R$ and mean accepted
length $\tau$ (Avg.\ over benchmarks and loads), greedy decoding. Strategies as
in Table~\ref{tab:tau-c1}. Best speedup per verifier and column in bold.}
\label{tab:main-speedup}
\centering
% Generated by tools/gen_experiment_tables.py -- do not edit; the caption and float live in sections/experiments.tex.

\footnotesize
\definecolor{famlin}{RGB}{249,245,252}\definecolor{famtrunc}{RGB}{254,251,229}\definecolor{famtree}{RGB}{240,249,242}\definecolor{famcad}{RGB}{236,244,252}\definecolor{famours}{RGB}{254,244,233}
\setlength{\tabcolsep}{2.0pt}
\renewcommand{\arraystretch}{1.12}
\begin{tabular}{lllcccccccccccccc}
\toprule
\multirow{2}{*}[-2.7pt]{Model} & \multirow{2}{*}[-2.7pt]{Strategy} & \multirow{2}{*}[-2.7pt]{Budget} & \multicolumn{7}{c}{Text benchmarks} & \multicolumn{7}{c}{Multimodal benchmarks} \\
\cmidrule(lr){4-10} \cmidrule(lr){11-17}
 & & & $R{=}1$ & $R{=}4$ & $R{=}8$ & $R{=}16$ & $R{=}32$ & $R{=}64$ & Avg.\ $\tau$ & $R{=}1$ & $R{=}4$ & $R{=}8$ & $R{=}16$ & $R{=}32$ & $R{=}64$ & Avg.\ $\tau$ \\
\midrule
\multirow{5}{*}{Q3.5-4B} & \cellcolor{famlin}DFlash & \cellcolor{famlin} & \cellcolor{famlin}3.35$\times$ & \cellcolor{famlin}3.33$\times$ & \cellcolor{famlin}2.92$\times$ & \cellcolor{famlin}1.92$\times$ & \cellcolor{famlin}1.31$\times$ & \cellcolor{famlin}1.02$\times$ & \cellcolor{famlin}5.36 & \cellcolor{famlin}2.54$\times$ & \cellcolor{famlin}2.39$\times$ & \cellcolor{famlin}2.09$\times$ & \cellcolor{famlin}1.40$\times$ & \cellcolor{famlin}1.01$\times$ & \cellcolor{famlin}0.78$\times$ & \cellcolor{famlin}4.06 \\
 & \cellcolor{famtrunc}DSpark & \cellcolor{famtrunc}\multirow{-2}{*}{$N{=}15$} & \cellcolor{famtrunc}3.19$\times$ & \cellcolor{famtrunc}3.31$\times$ & \cellcolor{famtrunc}3.05$\times$ & \cellcolor{famtrunc}2.42$\times$ & \cellcolor{famtrunc}1.78$\times$ & \cellcolor{famtrunc}1.40$\times$ & \cellcolor{famtrunc}4.95 & \cellcolor{famtrunc}2.35$\times$ & \cellcolor{famtrunc}2.34$\times$ & \cellcolor{famtrunc}2.18$\times$ & \cellcolor{famtrunc}1.82$\times$ & \cellcolor{famtrunc}1.41$\times$ & \cellcolor{famtrunc}1.17$\times$ & \cellcolor{famtrunc}3.66 \\
 & \cellcolor{famtree}DDTree & \cellcolor{famtree} & \cellcolor{famtree}3.64$\times$ & \cellcolor{famtree}3.49$\times$ & \cellcolor{famtree}2.89$\times$ & \cellcolor{famtree}1.83$\times$ & \cellcolor{famtree}1.20$\times$ & \cellcolor{famtree}0.94$\times$ & \cellcolor{famtree}6.16 & \cellcolor{famtree}2.78$\times$ & \cellcolor{famtree}2.55$\times$ & \cellcolor{famtree}2.13$\times$ & \cellcolor{famtree}1.40$\times$ & \cellcolor{famtree}0.97$\times$ & \cellcolor{famtree}0.75$\times$ & \cellcolor{famtree}4.76 \\
 & \cellcolor{famcad}CaDDTree & \cellcolor{famcad} & \cellcolor{famcad}2.77$\times$ & \cellcolor{famcad}2.72$\times$ & \cellcolor{famcad}2.44$\times$ & \cellcolor{famcad}1.86$\times$ & \cellcolor{famcad}1.30$\times$ & \cellcolor{famcad}1.03$\times$ & \cellcolor{famcad}6.16 & \cellcolor{famcad}2.11$\times$ & \cellcolor{famcad}2.05$\times$ & \cellcolor{famcad}1.82$\times$ & \cellcolor{famcad}1.39$\times$ & \cellcolor{famcad}1.03$\times$ & \cellcolor{famcad}0.83$\times$ & \cellcolor{famcad}4.75 \\
 & \cellcolor{famours}\textbf{D3Tree} & \cellcolor{famours}\multirow{-3}{*}{$B{=}15$} & \cellcolor{famours}\textbf{4.15}$\times$ & \cellcolor{famours}\textbf{3.78}$\times$ & \cellcolor{famours}\textbf{3.09}$\times$ & \cellcolor{famours}\textbf{2.60}$\times$ & \cellcolor{famours}\textbf{1.94}$\times$ & \cellcolor{famours}\textbf{1.56}$\times$ & \cellcolor{famours}5.57 & \cellcolor{famours}\textbf{3.06}$\times$ & \cellcolor{famours}\textbf{2.69}$\times$ & \cellcolor{famours}\textbf{2.34}$\times$ & \cellcolor{famours}\textbf{1.89}$\times$ & \cellcolor{famours}\textbf{1.48}$\times$ & \cellcolor{famours}\textbf{1.21}$\times$ & \cellcolor{famours}4.37 \\
\cmidrule(l){1-17}
\multirow{5}{*}{Q3.5-9B} & \cellcolor{famlin}DFlash & \cellcolor{famlin} & \cellcolor{famlin}3.92$\times$ & \cellcolor{famlin}3.39$\times$ & \cellcolor{famlin}2.61$\times$ & \cellcolor{famlin}1.88$\times$ & \cellcolor{famlin}1.36$\times$ & \cellcolor{famlin}1.00$\times$ & \cellcolor{famlin}5.24 & \cellcolor{famlin}2.89$\times$ & \cellcolor{famlin}2.24$\times$ & \cellcolor{famlin}1.82$\times$ & \cellcolor{famlin}1.38$\times$ & \cellcolor{famlin}1.04$\times$ & \cellcolor{famlin}0.77$\times$ & \cellcolor{famlin}3.93 \\
 & \cellcolor{famtrunc}DSpark & \cellcolor{famtrunc}\multirow{-2}{*}{$N{=}15$} & \cellcolor{famtrunc}3.53$\times$ & \cellcolor{famtrunc}3.27$\times$ & \cellcolor{famtrunc}2.90$\times$ & \cellcolor{famtrunc}2.33$\times$ & \cellcolor{famtrunc}1.73$\times$ & \cellcolor{famtrunc}1.42$\times$ & \cellcolor{famtrunc}4.70 & \cellcolor{famtrunc}2.58$\times$ & \cellcolor{famtrunc}2.36$\times$ & \cellcolor{famtrunc}2.09$\times$ & \cellcolor{famtrunc}1.73$\times$ & \cellcolor{famtrunc}\textbf{1.41}$\times$ & \cellcolor{famtrunc}\textbf{1.14}$\times$ & \cellcolor{famtrunc}3.52 \\
 & \cellcolor{famtree}DDTree & \cellcolor{famtree} & \cellcolor{famtree}4.31$\times$ & \cellcolor{famtree}3.50$\times$ & \cellcolor{famtree}2.80$\times$ & \cellcolor{famtree}1.82$\times$ & \cellcolor{famtree}1.27$\times$ & \cellcolor{famtree}0.93$\times$ & \cellcolor{famtree}6.05 & \cellcolor{famtree}3.22$\times$ & \cellcolor{famtree}2.51$\times$ & \cellcolor{famtree}2.01$\times$ & \cellcolor{famtree}1.35$\times$ & \cellcolor{famtree}0.96$\times$ & \cellcolor{famtree}0.72$\times$ & \cellcolor{famtree}4.63 \\
 & \cellcolor{famcad}CaDDTree & \cellcolor{famcad} & \cellcolor{famcad}3.55$\times$ & \cellcolor{famcad}3.06$\times$ & \cellcolor{famcad}2.71$\times$ & \cellcolor{famcad}1.87$\times$ & \cellcolor{famcad}1.39$\times$ & \cellcolor{famcad}1.04$\times$ & \cellcolor{famcad}6.06 & \cellcolor{famcad}2.65$\times$ & \cellcolor{famcad}2.23$\times$ & \cellcolor{famcad}1.91$\times$ & \cellcolor{famcad}1.37$\times$ & \cellcolor{famcad}1.04$\times$ & \cellcolor{famcad}0.79$\times$ & \cellcolor{famcad}4.61 \\
 & \cellcolor{famours}\textbf{D3Tree} & \cellcolor{famours}\multirow{-3}{*}{$B{=}15$} & \cellcolor{famours}\textbf{4.40}$\times$ & \cellcolor{famours}\textbf{3.65}$\times$ & \cellcolor{famours}\textbf{3.04}$\times$ & \cellcolor{famours}\textbf{2.53}$\times$ & \cellcolor{famours}\textbf{1.84}$\times$ & \cellcolor{famours}\textbf{1.46}$\times$ & \cellcolor{famours}5.45 & \cellcolor{famours}\textbf{3.26}$\times$ & \cellcolor{famours}\textbf{2.62}$\times$ & \cellcolor{famours}\textbf{2.24}$\times$ & \cellcolor{famours}\textbf{1.88}$\times$ & \cellcolor{famours}1.40$\times$ & \cellcolor{famours}1.13$\times$ & \cellcolor{famours}4.24 \\
\cmidrule(l){1-17}
\multirow{5}{*}{IVL3.5-4B} & \cellcolor{famlin}DFlash & \cellcolor{famlin} & \cellcolor{famlin}3.56$\times$ & \cellcolor{famlin}3.23$\times$ & \cellcolor{famlin}2.66$\times$ & \cellcolor{famlin}1.67$\times$ & \cellcolor{famlin}1.24$\times$ & \cellcolor{famlin}0.95$\times$ & \cellcolor{famlin}4.69 & \cellcolor{famlin}2.37$\times$ & \cellcolor{famlin}2.04$\times$ & \cellcolor{famlin}1.68$\times$ & \cellcolor{famlin}1.24$\times$ & \cellcolor{famlin}1.09$\times$ & \cellcolor{famlin}0.93$\times$ & \cellcolor{famlin}3.70 \\
 & \cellcolor{famtrunc}DSpark & \cellcolor{famtrunc}\multirow{-2}{*}{$N{=}15$} & \cellcolor{famtrunc}3.17$\times$ & \cellcolor{famtrunc}3.09$\times$ & \cellcolor{famtrunc}2.83$\times$ & \cellcolor{famtrunc}2.19$\times$ & \cellcolor{famtrunc}1.61$\times$ & \cellcolor{famtrunc}1.31$\times$ & \cellcolor{famtrunc}4.32 & \cellcolor{famtrunc}2.05$\times$ & \cellcolor{famtrunc}1.90$\times$ & \cellcolor{famtrunc}1.70$\times$ & \cellcolor{famtrunc}1.52$\times$ & \cellcolor{famtrunc}1.29$\times$ & \cellcolor{famtrunc}1.21$\times$ & \cellcolor{famtrunc}3.29 \\
 & \cellcolor{famtree}DDTree & \cellcolor{famtree} & \cellcolor{famtree}3.64$\times$ & \cellcolor{famtree}3.02$\times$ & \cellcolor{famtree}2.16$\times$ & \cellcolor{famtree}1.19$\times$ & \cellcolor{famtree}0.77$\times$ & \cellcolor{famtree}0.63$\times$ & \cellcolor{famtree}5.44 & \cellcolor{famtree}2.36$\times$ & \cellcolor{famtree}1.89$\times$ & \cellcolor{famtree}1.45$\times$ & \cellcolor{famtree}1.01$\times$ & \cellcolor{famtree}0.77$\times$ & \cellcolor{famtree}0.64$\times$ & \cellcolor{famtree}4.36 \\
 & \cellcolor{famcad}CaDDTree & \cellcolor{famcad} & \cellcolor{famcad}2.70$\times$ & \cellcolor{famcad}2.56$\times$ & \cellcolor{famcad}2.36$\times$ & \cellcolor{famcad}1.73$\times$ & \cellcolor{famcad}1.36$\times$ & \cellcolor{famcad}1.03$\times$ & \cellcolor{famcad}5.44 & \cellcolor{famcad}1.92$\times$ & \cellcolor{famcad}1.75$\times$ & \cellcolor{famcad}1.51$\times$ & \cellcolor{famcad}1.26$\times$ & \cellcolor{famcad}1.13$\times$ & \cellcolor{famcad}0.99$\times$ & \cellcolor{famcad}4.36 \\
 & \cellcolor{famours}\textbf{D3Tree} & \cellcolor{famours}\multirow{-3}{*}{$B{=}15$} & \cellcolor{famours}\textbf{3.95}$\times$ & \cellcolor{famours}\textbf{3.63}$\times$ & \cellcolor{famours}\textbf{3.10}$\times$ & \cellcolor{famours}\textbf{2.45}$\times$ & \cellcolor{famours}\textbf{1.83}$\times$ & \cellcolor{famours}\textbf{1.45}$\times$ & \cellcolor{famours}5.07 & \cellcolor{famours}\textbf{2.54}$\times$ & \cellcolor{famours}\textbf{2.24}$\times$ & \cellcolor{famours}\textbf{1.86}$\times$ & \cellcolor{famours}\textbf{1.63}$\times$ & \cellcolor{famours}\textbf{1.41}$\times$ & \cellcolor{famours}\textbf{1.28}$\times$ & \cellcolor{famours}4.16 \\
\cmidrule(l){1-17}
\multirow{5}{*}{IVL3.5-8B} & \cellcolor{famlin}DFlash & \cellcolor{famlin} & \cellcolor{famlin}4.17$\times$ & \cellcolor{famlin}3.57$\times$ & \cellcolor{famlin}3.05$\times$ & \cellcolor{famlin}2.09$\times$ & \cellcolor{famlin}1.60$\times$ & \cellcolor{famlin}1.17$\times$ & \cellcolor{famlin}5.39 & \cellcolor{famlin}2.54$\times$ & \cellcolor{famlin}2.12$\times$ & \cellcolor{famlin}1.86$\times$ & \cellcolor{famlin}1.36$\times$ & \cellcolor{famlin}1.19$\times$ & \cellcolor{famlin}0.97$\times$ & \cellcolor{famlin}3.69 \\
 & \cellcolor{famtrunc}DSpark & \cellcolor{famtrunc}\multirow{-2}{*}{$N{=}15$} & \cellcolor{famtrunc}3.88$\times$ & \cellcolor{famtrunc}3.47$\times$ & \cellcolor{famtrunc}3.09$\times$ & \cellcolor{famtrunc}2.51$\times$ & \cellcolor{famtrunc}1.87$\times$ & \cellcolor{famtrunc}1.45$\times$ & \cellcolor{famtrunc}5.12 & \cellcolor{famtrunc}2.24$\times$ & \cellcolor{famtrunc}2.06$\times$ & \cellcolor{famtrunc}1.87$\times$ & \cellcolor{famtrunc}1.63$\times$ & \cellcolor{famtrunc}1.41$\times$ & \cellcolor{famtrunc}1.23$\times$ & \cellcolor{famtrunc}3.35 \\
 & \cellcolor{famtree}DDTree & \cellcolor{famtree} & \cellcolor{famtree}4.51$\times$ & \cellcolor{famtree}3.67$\times$ & \cellcolor{famtree}2.77$\times$ & \cellcolor{famtree}1.58$\times$ & \cellcolor{famtree}1.04$\times$ & \cellcolor{famtree}0.79$\times$ & \cellcolor{famtree}6.11 & \cellcolor{famtree}2.76$\times$ & \cellcolor{famtree}2.10$\times$ & \cellcolor{famtree}1.67$\times$ & \cellcolor{famtree}1.14$\times$ & \cellcolor{famtree}0.87$\times$ & \cellcolor{famtree}0.74$\times$ & \cellcolor{famtree}4.33 \\
 & \cellcolor{famcad}CaDDTree & \cellcolor{famcad} & \cellcolor{famcad}3.57$\times$ & \cellcolor{famcad}3.01$\times$ & \cellcolor{famcad}2.70$\times$ & \cellcolor{famcad}1.93$\times$ & \cellcolor{famcad}1.56$\times$ & \cellcolor{famcad}1.16$\times$ & \cellcolor{famcad}6.09 & \cellcolor{famcad}2.23$\times$ & \cellcolor{famcad}1.93$\times$ & \cellcolor{famcad}1.65$\times$ & \cellcolor{famcad}1.30$\times$ & \cellcolor{famcad}1.15$\times$ & \cellcolor{famcad}1.03$\times$ & \cellcolor{famcad}4.33 \\
 & \cellcolor{famours}\textbf{D3Tree} & \cellcolor{famours}\multirow{-3}{*}{$B{=}15$} & \cellcolor{famours}\textbf{4.70}$\times$ & \cellcolor{famours}\textbf{3.97}$\times$ & \cellcolor{famours}\textbf{3.42}$\times$ & \cellcolor{famours}\textbf{2.74}$\times$ & \cellcolor{famours}\textbf{1.99}$\times$ & \cellcolor{famours}\textbf{1.59}$\times$ & \cellcolor{famours}5.70 & \cellcolor{famours}\textbf{2.87}$\times$ & \cellcolor{famours}\textbf{2.38}$\times$ & \cellcolor{famours}\textbf{2.02}$\times$ & \cellcolor{famours}\textbf{1.74}$\times$ & \cellcolor{famours}\textbf{1.44}$\times$ & \cellcolor{famours}\textbf{1.32}$\times$ & \cellcolor{famours}4.16 \\
\cmidrule(l){1-17}
\multirow{5}{*}{G4-E2B} & \cellcolor{famlin}DFlash & \cellcolor{famlin} & \cellcolor{famlin}5.10$\times$ & \cellcolor{famlin}4.88$\times$ & \cellcolor{famlin}3.67$\times$ & \cellcolor{famlin}2.56$\times$ & \cellcolor{famlin}1.71$\times$ & \cellcolor{famlin}1.32$\times$ & \cellcolor{famlin}5.33 & \cellcolor{famlin}2.84$\times$ & \cellcolor{famlin}2.48$\times$ & \cellcolor{famlin}1.95$\times$ & \cellcolor{famlin}1.46$\times$ & \cellcolor{famlin}1.11$\times$ & \cellcolor{famlin}0.92$\times$ & \cellcolor{famlin}3.23 \\
 & \cellcolor{famtrunc}DSpark & \cellcolor{famtrunc}\multirow{-2}{*}{$N{=}15$} & \cellcolor{famtrunc}4.61$\times$ & \cellcolor{famtrunc}4.52$\times$ & \cellcolor{famtrunc}3.71$\times$ & \cellcolor{famtrunc}2.86$\times$ & \cellcolor{famtrunc}2.09$\times$ & \cellcolor{famtrunc}1.63$\times$ & \cellcolor{famtrunc}4.93 & \cellcolor{famtrunc}2.33$\times$ & \cellcolor{famtrunc}2.18$\times$ & \cellcolor{famtrunc}1.94$\times$ & \cellcolor{famtrunc}1.64$\times$ & \cellcolor{famtrunc}1.34$\times$ & \cellcolor{famtrunc}1.21$\times$ & \cellcolor{famtrunc}2.88 \\
 & \cellcolor{famtree}DDTree & \cellcolor{famtree} & \cellcolor{famtree}5.24$\times$ & \cellcolor{famtree}3.51$\times$ & \cellcolor{famtree}1.82$\times$ & \cellcolor{famtree}0.79$\times$ & \cellcolor{famtree}0.51$\times$ & \cellcolor{famtree}0.38$\times$ & \cellcolor{famtree}6.13 & \cellcolor{famtree}2.68$\times$ & \cellcolor{famtree}1.79$\times$ & \cellcolor{famtree}1.03$\times$ & \cellcolor{famtree}0.49$\times$ & \cellcolor{famtree}0.28$\times$ & \cellcolor{famtree}0.25$\times$ & \cellcolor{famtree}3.91 \\
 & \cellcolor{famcad}CaDDTree & \cellcolor{famcad} & \cellcolor{famcad}3.54$\times$ & \cellcolor{famcad}3.41$\times$ & \cellcolor{famcad}2.79$\times$ & \cellcolor{famcad}2.31$\times$ & \cellcolor{famcad}1.59$\times$ & \cellcolor{famcad}1.35$\times$ & \cellcolor{famcad}6.08 & \cellcolor{famcad}2.13$\times$ & \cellcolor{famcad}1.95$\times$ & \cellcolor{famcad}1.74$\times$ & \cellcolor{famcad}1.42$\times$ & \cellcolor{famcad}1.15$\times$ & \cellcolor{famcad}1.07$\times$ & \cellcolor{famcad}3.90 \\
 & \cellcolor{famours}\textbf{D3Tree} & \cellcolor{famours}\multirow{-3}{*}{$B{=}15$} & \cellcolor{famours}\textbf{5.42}$\times$ & \cellcolor{famours}\textbf{5.12}$\times$ & \cellcolor{famours}\textbf{4.12}$\times$ & \cellcolor{famours}\textbf{3.32}$\times$ & \cellcolor{famours}\textbf{2.61}$\times$ & \cellcolor{famours}\textbf{2.07}$\times$ & \cellcolor{famours}5.69 & \cellcolor{famours}\textbf{2.95}$\times$ & \cellcolor{famours}\textbf{2.76}$\times$ & \cellcolor{famours}\textbf{2.33}$\times$ & \cellcolor{famours}\textbf{1.98}$\times$ & \cellcolor{famours}\textbf{1.67}$\times$ & \cellcolor{famours}\textbf{1.52}$\times$ & \cellcolor{famours}3.70 \\
\cmidrule(l){1-17}
\multirow{5}{*}{G4-E4B} & \cellcolor{famlin}DFlash & \cellcolor{famlin} & \cellcolor{famlin}4.77$\times$ & \cellcolor{famlin}4.25$\times$ & \cellcolor{famlin}3.19$\times$ & \cellcolor{famlin}2.29$\times$ & \cellcolor{famlin}1.47$\times$ & \cellcolor{famlin}0.93$\times$ & \cellcolor{famlin}5.83 & \cellcolor{famlin}2.53$\times$ & \cellcolor{famlin}2.19$\times$ & \cellcolor{famlin}1.75$\times$ & \cellcolor{famlin}1.33$\times$ & \cellcolor{famlin}1.02$\times$ & \cellcolor{famlin}0.84$\times$ & \cellcolor{famlin}3.43 \\
 & \cellcolor{famtrunc}DSpark & \cellcolor{famtrunc}\multirow{-2}{*}{$N{=}15$} & \cellcolor{famtrunc}4.38$\times$ & \cellcolor{famtrunc}3.94$\times$ & \cellcolor{famtrunc}3.09$\times$ & \cellcolor{famtrunc}2.45$\times$ & \cellcolor{famtrunc}1.75$\times$ & \cellcolor{famtrunc}1.15$\times$ & \cellcolor{famtrunc}5.46 & \cellcolor{famtrunc}2.20$\times$ & \cellcolor{famtrunc}1.95$\times$ & \cellcolor{famtrunc}1.74$\times$ & \cellcolor{famtrunc}1.51$\times$ & \cellcolor{famtrunc}1.28$\times$ & \cellcolor{famtrunc}1.12$\times$ & \cellcolor{famtrunc}3.00 \\
 & \cellcolor{famtree}DDTree & \cellcolor{famtree} & \cellcolor{famtree}5.02$\times$ & \cellcolor{famtree}3.00$\times$ & \cellcolor{famtree}1.56$\times$ & \cellcolor{famtree}0.70$\times$ & \cellcolor{famtree}0.43$\times$ & \cellcolor{famtree}0.34$\times$ & \cellcolor{famtree}6.56 & \cellcolor{famtree}2.74$\times$ & \cellcolor{famtree}1.56$\times$ & \cellcolor{famtree}0.84$\times$ & \cellcolor{famtree}0.42$\times$ & \cellcolor{famtree}0.21$\times$ & \cellcolor{famtree}0.17$\times$ & \cellcolor{famtree}4.12 \\
 & \cellcolor{famcad}CaDDTree & \cellcolor{famcad} & \cellcolor{famcad}3.45$\times$ & \cellcolor{famcad}3.32$\times$ & \cellcolor{famcad}2.70$\times$ & \cellcolor{famcad}2.12$\times$ & \cellcolor{famcad}1.50$\times$ & \cellcolor{famcad}1.22$\times$ & \cellcolor{famcad}6.55 & \cellcolor{famcad}1.97$\times$ & \cellcolor{famcad}1.82$\times$ & \cellcolor{famcad}1.56$\times$ & \cellcolor{famcad}1.28$\times$ & \cellcolor{famcad}1.02$\times$ & \cellcolor{famcad}0.85$\times$ & \cellcolor{famcad}4.12 \\
 & \cellcolor{famours}\textbf{D3Tree} & \cellcolor{famours}\multirow{-3}{*}{$B{=}15$} & \cellcolor{famours}\textbf{5.21}$\times$ & \cellcolor{famours}\textbf{4.58}$\times$ & \cellcolor{famours}\textbf{3.60}$\times$ & \cellcolor{famours}\textbf{2.86}$\times$ & \cellcolor{famours}\textbf{2.08}$\times$ & \cellcolor{famours}\textbf{1.63}$\times$ & \cellcolor{famours}6.10 & \cellcolor{famours}\textbf{2.83}$\times$ & \cellcolor{famours}\textbf{2.39}$\times$ & \cellcolor{famours}\textbf{1.93}$\times$ & \cellcolor{famours}\textbf{1.66}$\times$ & \cellcolor{famours}\textbf{1.39}$\times$ & \cellcolor{famours}\textbf{1.14}$\times$ & \cellcolor{famours}3.90 \\
\bottomrule
\end{tabular}

\end{table*}

At inference the embeddings and head weights are fixed, so
\eqref{eq:survival-logit} factors into a vocabulary table and a per-position
term:
\begin{align}
 \nu_a&=w_e^\top e_a\quad(a\in\mathcal{V}_T),\qquad
 o_d=w_h^\top h_d+b,\label{eq:cached-projection}\\
 q_{d,a}&=\sigma\bigl(o_d+\nu_a+\alpha\log p_d(a)\bigr).
 \label{eq:fused-head}
\end{align}
The table is built once, after the drafter shares the target embeddings, and
is invalidated whenever weights are reloaded or replaced. An optional fused
GPU kernel then computes the position term, the table lookup, the
log-probability term, and the sigmoid without materializing a
$D\times K\times H$ tensor of candidate embeddings, following the IO-aware
design of fused attention
kernels~\cite{dao2022flashattention,dao2024flashattention2}. For $R$ requests, the
head arithmetic changes from
\begin{equation}
 \underbrace{O(RDKH)}_{\text{candidate projections}}
 \quad\longrightarrow\quad
 \underbrace{O(RDH+RDK)}_{\text{per round}},
 \label{eq:head-complexity}
\end{equation}
with one-time $O(|\mathcal{V}_T|H)$ projection work and
$O(|\mathcal{V}_T|)$ cached scalars. This count excludes draft-logit
generation, tree construction, and target verification. The fused path
requires a single tensor-parallel shard and unquantized FP16/BF16
embeddings without added vocabulary; other configurations use the direct
head. The factorization is exact only in exact arithmetic: reordered
low-precision reductions can change scores slightly and, near the threshold,
the resulting pruning decisions.

\subsection{Overlapped Execution}
\label{sec:overlap}
A decoding round of D3Tree stays on the GPU. Candidate extraction,
construction (Section~\ref{sec:candidate-tree}), survival scoring, the
batch-wide threshold~\eqref{eq:step-threshold}, and pruning run there, and
only the retained size of each request is copied to the host,
asynchronously. The host can therefore prepare step $t{+}1$ while the GPU
executes step $t$, except for two dependencies. First, the inputs of step
$t{+}1$ depend on the tokens accepted in step $t$. The request state kept on
the GPU is authoritative: the host prepares the next step without waiting
for the accepted counts and commits its own copy behind the queued work;
only steps that finish or preempt requests synchronize. Second, the width of
step $t{+}1$ depends on the retained sizes $N_i$, which exist only after
drafting. The host lays out every request at its reserved width $B{+}1$,
and the GPU trims the batch to the retained trees before the target pass.
Verification batches of up to 1024 query rows (512 on hybrid targets) run as
captured full CUDA graphs; on hybrid targets the shared node-state scratch
also limits this trimming to 256 reserved rows. Steps beyond these limits, or
that admit new requests or encoder inputs, wait for the retained sizes
instead.

\section{Experiments}
\label{sec:experiments}
% Numbers: results/d3tree_main_20261106.json (tools/extract_d3m_results.py on every node);
% tables: tools/gen_experiment_tables.py, which also prints the figures quoted below.
% Raw runs: Speculator_pre5/output/d3tree_main_20261106 (README.md) and the per-node copies.
We evaluate D3Tree as a verification strategy for a fixed block-diffusion
drafter: every strategy for a verifier uses the same drafter, prompts,
generation limits, and serving runtime, and only the set of draft tokens sent
to the target differs. Section~\ref{sec:main-results} compares D3Tree with
linear, dynamically truncated, and fixed-budget tree verification across six
verifiers and six loads, and Section~\ref{sec:caddtree-results} compares it
with cost-selected trees. Section~\ref{sec:load-trace} follows the
allocation within a run, at a fixed concurrency and under a time-varying
arrival trace, Section~\ref{sec:ablation} varies the construction budget,
and Section~\ref{sec:device-results} repeats the comparison on other GPUs.

% Floats placed ahead of their sections so that, at one figure or table per page, none lands among the references.
\begin{figure*}[t]
\centering
\includegraphics[width=\linewidth]{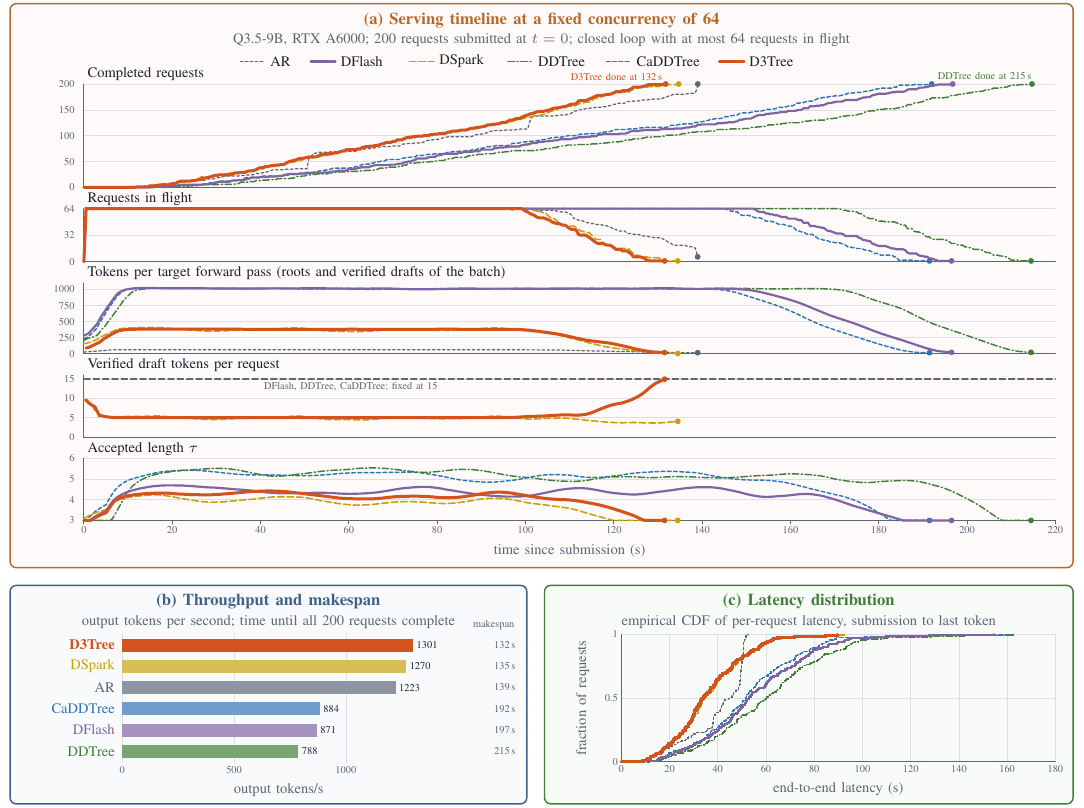}
\caption{\normalfont Every strategy serving 200 requests at a fixed
concurrency of 64 (Q3.5-9B, RTX A6000; all requests are due at once, as in
the sweeps of Table~\ref{tab:main-speedup}). (a)~Each curve ends, with a
dot, when the last request of that run finishes. From top: requests
completed; requests in flight; tokens per target pass, i.e., the roots and
verified draft tokens of all decoding requests; verified draft tokens per
request and decoding round; and accepted length $\tau$ (sliding means over
about 12, 12, and 28\,s). (b)~Output throughput over the run and the time
until all 200 requests have finished. (c)~Distribution of the latency from
sending a request to its last token.}
\label{fig:closed-loop}
\end{figure*}

\begin{figure*}[t]
\centering
\includegraphics[width=\linewidth]{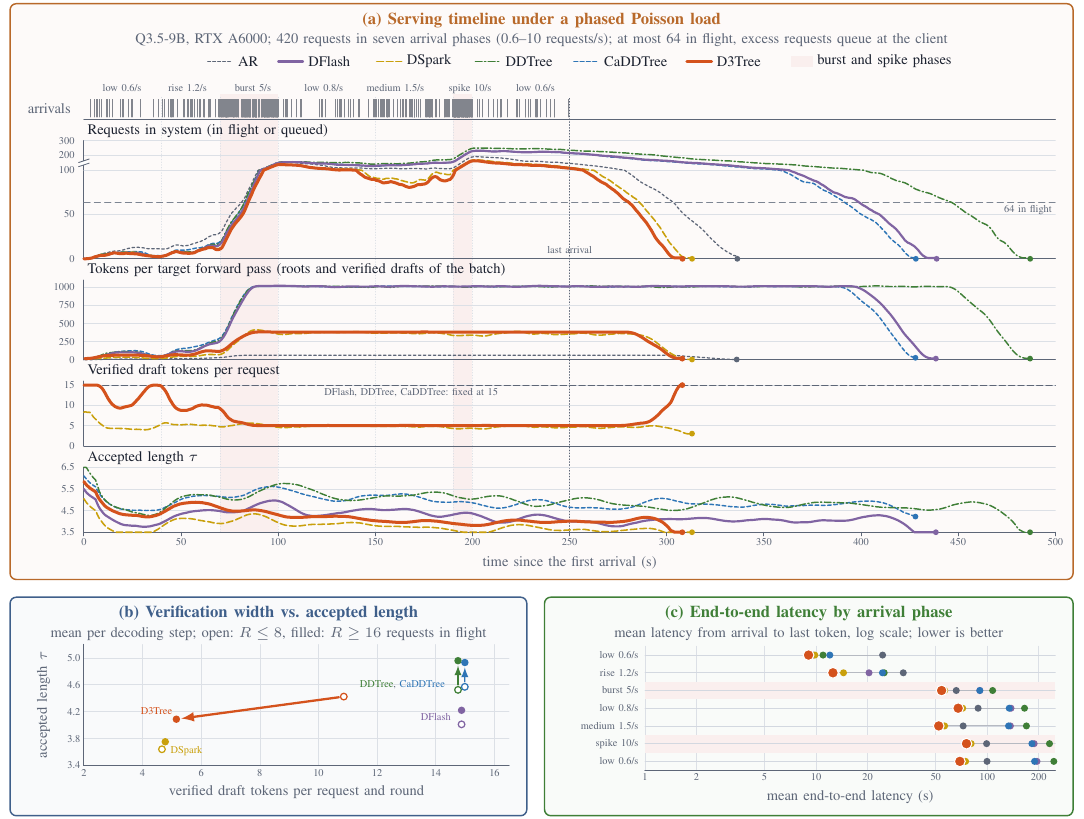}
\caption{\normalfont Every strategy serving the same time-varying load.
(a)~Each curve ends, with a dot, when the last request of that run
finishes. From top: one tick per arriving request; requests in the system
(in flight or waiting at the client; 4-s moving average; the axis is
compressed above 100); and, as in Fig.~\ref{fig:closed-loop}, tokens per
target pass, verified draft tokens per request and decoding round, and
accepted length $\tau$. (b)~Mean verified draft tokens and accepted length
$\tau$ over the decoding steps with at most 8 (open) and at least 16
(filled) requests in flight; AR is omitted. (c)~Mean end-to-end latency,
from arrival to the last token, of the requests arriving in each phase
(log scale).}
\label{fig:load-trace}
\end{figure*}

\subsection{Experimental Setup}
\label{sec:setup}

\textbf{Verifiers and workloads.}
We use six frozen vision-language verifiers: Qwen3.5-4B/9B (Q3.5),
InternVL3.5-4B/8B (IVL3.5), and Gemma~4 E2B/E4B (G4). The Qwen3.5 models are
hybrid targets with gated DeltaNet layers. The seven text benchmarks cover
mathematics, code, and dialogue: GSM8K~\cite{cobbe2021gsm8k},
MATH-500~\cite{hendrycks2021math,lightman2023math500}, AIME25,
HumanEval~\cite{chen2021humaneval}, MBPP~\cite{austin2021mbpp},
LiveCodeBench~\cite{jain2024livecodebench}, and
MT-Bench~\cite{zheng2023mtbench}. The ten multimodal benchmarks cover
documents, charts, visual reasoning, captioning, and dialogue:
OmniDocBench~\cite{ouyang2024omnidocbench},
OCRBench-v2~\cite{fu2024ocrbenchv2}, DocVQA~\cite{mathew2020docvqa},
olmOCR~\cite{poznanski2025olmocr}, CharXiv~\cite{wang2024charxiv},
TextVQA~\cite{singh2019textvqa}, MMMU-Pro~\cite{yue2024mmmupro},
GQA~\cite{hudson2019gqa}, COCO Caption~\cite{chen2015cococaptions}, and
MM-MT-Bench~\cite{agrawal2024pixtral}.

\textbf{Drafter and heads.}
Each verifier has a six-layer block-diffusion drafter with the DFlash
architecture~\cite{chen2026dflash}, a compressed 32{,}768-token vocabulary,
and a draft horizon of $D=15$ positions. The drafters are trained on one
million prompts whose responses are regenerated by the corresponding
verifier, and their weights are frozen in all experiments. D3Tree serves
with the configuration of Section~\ref{sec:admission}: $B=15$,
whole trees up to $R_{\rm full}$ requests, and the tile-aligned batch-wide
node budget beyond it.

\textbf{Baselines.}
\emph{DFlash}~\cite{chen2026dflash} verifies the full 15-token linear
draft. \emph{DSpark}~\cite{cheng2026dspark} verifies a prefix whose length
is chosen per request by the drafter's learned length head with the chain
rule~\eqref{eq:chain-length} and threshold $0.3$; it reproduces DSpark's
learned length selection, not its serving scheduler.
\emph{DDTree}~\cite{ringel2026ddtree} verifies the full best-first tree of
$B\in\{15,31,63\}$ nodes. It uses the same tree builder as D3Tree, so its
trees at $B=15$ are exactly those that D3Tree prunes. Tables~\ref{tab:tau-c1}--\ref{tab:main-speedup}
report $B=15$, the construction budget of D3Tree.
\emph{CaDDTree}~\cite{zhang2026caddtree} is our reimplementation of
cost-selected trees on the same runtime. For each request and round it
grows the best-first tree under the score~\eqref{eq:path-score} while the
estimated throughput $\bigl(1+\sum_{v}e^{\ell(v)}\bigr)/T(n;c)$ increases,
where the sum runs over the first $n$ nodes in best-first order, $\ell(v)$ is
the path score~\eqref{eq:path-score}, and $T(n;c)$ is the single-request
round latency for $n$ nodes at context length $c$. We profile $T$ on every
verifier and GPU at $n\in\{0,15,\ldots,127\}$ and $c$ from 1K to 16K tokens and fit it
monotone and convex, following the original method. We run budgets
$B\in\{15,31,63,127\}$.

\textbf{Protocol and metrics.}
Each benchmark supplies at most 100 prompts, outputs are capped at 1024
tokens with the native end-of-sequence token, and decoding is greedy.
Fixed-load sweeps use client concurrency $R\in\{1,4,8,16,32,64\}$, with
three warm-up requests before every measurement. We report the accepted
length $\tau=\mathbb{E}[A]+1$, the mean number of tokens committed per
verified draft including the bonus token, and output throughput over the
elapsed benchmark time, including prefill. The speedup of a benchmark group
divides the unweighted mean throughput over its benchmarks by the same mean
of autoregressive decoding (AR) for the same verifier, GPU, load, and
prompts. Avg.\ $\tau$ averages over the benchmarks of a group and the six
loads.

\textbf{Hardware and runtime.}
All methods, AR included, run on one build of vLLM~\cite{kwon2023vllm}
with CUDA graphs and at most 64 concurrent sequences; speculative methods
use the FlexAttention~\cite{dong2024flex} backend, with tree rows on the
Triton tree-attention kernel of Section~\ref{sec:verification}. Q3.5-4B, IVL3.5-4B, and G4-E2B run on RTX
6000 Ada GPUs (48\,GB), and Q3.5-9B, IVL3.5-8B, and G4-E4B on RTX A6000
GPUs (48\,GB). All strategies of one verifier run on the same machine, so
every speedup compares measurements on the same GPU type.
Section~\ref{sec:device-results} repeats the comparison for Q3.5-4B,
IVL3.5-4B, and G4-E2B on an RTX 5090 (32\,GB) and for all six verifiers on an
A800 (80\,GB).

\subsection{Main Results}
\label{sec:main-results}

\textbf{Single request.}
With few concurrent requests, the target pass is bound by reading the
weights, and D3Tree verifies whole trees. Tables~\ref{tab:tau-c1}
and~\ref{tab:speedup-c1} report the accepted length and speedup of every
strategy per benchmark at $R=1$. Table~\ref{tab:caddtree}
confirms that at $R=1$ it verifies all 15 nodes per tree and matches the
accepted length of DDTree $B=15$ (for Q3.5-4B, 6.16 vs.\ 6.18 tokens on
text and 4.75 vs.\ 4.76 on multimodal benchmarks). Its single-request
speedup over all 17 benchmarks is 3.17--4.03$\times$, up to 12\% above
DDTree $B=15$ on the same trees; this margin comes from the GPU-resident
construction and the overlapped execution of Section~\ref{sec:overlap}, not
from pruning. Across the 102 verifier--benchmark entries, the
accepted length of D3Tree differs from that of DDTree $B=15$ by 0.05 tokens
on average, and exceeds that of DFlash's linear draft by 0.28--0.99 tokens.
D3Tree is the fastest strategy on 91 entries, with speedups of
1.60--6.60$\times$ over AR, and is faster than DFlash on 95. Eight of the
eleven exceptions are G4-E2B, the smallest verifier, on which DFlash is
ahead on six benchmarks by up to 0.17$\times$; on the other three, DDTree
$B=15$ is ahead by at most 0.03$\times$.
Larger DDTree budgets raise the accepted length further
($B=63$ reaches 7.0 tokens on text for Q3.5-4B), but they are faster than
D3Tree in at most four settings at $R=1$, by 0.01--0.11$\times$, and fall
further behind under load.

\textbf{Concurrent requests.}
Table~\ref{tab:main-speedup} reports the speedup of every strategy at all
six loads. D3Tree attains the highest speedup in 70 of the 72
verifier--group--load settings; the two exceptions are DSpark on the
multimodal benchmarks of Q3.5-9B at $R=32$ and $R=64$, each ahead by
0.01$\times$. As the load grows, the cost of a target pass follows the number of
verified rows, and fixed trees lose most of their advantage: DDTree falls
below AR at $R=64$ on every verifier (0.26--0.84$\times$ over all
benchmarks). DSpark, which shortens its chains, is the strongest baseline
here. D3Tree prunes once the load exceeds $R_{\rm full}$: at
$R=8$ for the hybrid Qwen3.5 targets and at $R=16$ for the others, it keeps
7.4--8.5 nodes per tree, and 5.8--7.0 at $R\geq32$
(Table~\ref{tab:caddtree}). Pruning more than halves the verified
rows but costs only 0.6--0.8 accepted tokens, because the batch-wide budget
keeps the highest-scoring nodes. At $R=64$, D3Tree reaches
1.28--1.81$\times$ over all benchmarks, compared with 1.14--1.44$\times$ for
the best baseline of each verifier, and 1.36--1.59$\times$ the throughput of
DFlash's linear verification of the same drafts.

\subsection{Comparison with Cost-Selected Trees}
\label{sec:caddtree-results}

\begin{table*}[p]
\caption{\normalfont Every strategy at every verification budget over all
17 benchmarks: speedup over AR, and accepted length $\tau$ / verified draft
tokens per request and decoding round (tree nodes, root excluded), at each
concurrency $R$. DFlash verifies the first $N$ tokens of the linear draft
(budget $N$); DSpark a prefix of at most $N=15$ tokens chosen by the
drafter's length head; DDTree verifies the full best-first tree of $B$
nodes, CaDDTree a cost-selected prefix of its best-first tree, and D3Tree
prunes the $B$-node tree under its load-dependent node budget. Best speedup per verifier and
column in bold.}
\label{tab:caddtree}
\centering
% Generated by tools/gen_experiment_tables.py -- do not edit; the caption and float live in sections/experiments.tex.

\scriptsize
\definecolor{famlin}{RGB}{249,245,252}\definecolor{famtrunc}{RGB}{254,251,229}\definecolor{famtree}{RGB}{240,249,242}\definecolor{famcad}{RGB}{236,244,252}\definecolor{famours}{RGB}{254,244,233}
\ifdefined\taucw\else\newlength\taucw\newlength\tauncw\fi
\settowidth\taucw{0.00}\settowidth\tauncw{000.0}
\providecommand\taun[2]{\mbox{\makebox[\taucw][r]{#1}\ /\ \makebox[\tauncw][l]{#2}}}
\setlength{\tabcolsep}{2.6pt}
\renewcommand{\arraystretch}{0.94}
\setlength{\aboverulesep}{0.15ex}
\setlength{\belowrulesep}{0.25ex}
\begin{tabularx}{\textwidth}{lll>{\centering\arraybackslash}X>{\centering\arraybackslash}X>{\centering\arraybackslash}X>{\centering\arraybackslash}X>{\centering\arraybackslash}X>{\centering\arraybackslash}Xcccccc}
\toprule
\multirow{2}{*}[-2.7pt]{Model} & \multirow{2}{*}[-2.7pt]{Strategy} & \multirow{2}{*}[-2.7pt]{Budget} & \multicolumn{6}{c}{Speedup over AR at concurrency $R$} & \multicolumn{6}{c}{Accepted length $\tau$\,/\,verified draft tokens at concurrency $R$} \\
\cmidrule(lr){4-9} \cmidrule(lr){10-15}
 & & & 1 & 4 & 8 & 16 & 32 & 64 & 1 & 4 & 8 & 16 & 32 & 64 \\
\midrule
\multirow{12}{*}{Q3.5-4B} & \cellcolor{famlin} & \cellcolor{famlin}$N{=}5$ & \cellcolor{famlin}2.44$\times$ & \cellcolor{famlin}2.41$\times$ & \cellcolor{famlin}2.34$\times$ & \cellcolor{famlin}2.00$\times$ & \cellcolor{famlin}1.58$\times$ & \cellcolor{famlin}1.29$\times$ & \cellcolor{famlin}\taun{3.67}{5.0} & \cellcolor{famlin}\taun{3.67}{5.0} & \cellcolor{famlin}\taun{3.67}{5.0} & \cellcolor{famlin}\taun{3.66}{5.0} & \cellcolor{famlin}\taun{3.66}{5.0} & \cellcolor{famlin}\taun{3.67}{5.0} \\
 & \cellcolor{famlin} & \cellcolor{famlin}$N{=}7$ & \cellcolor{famlin}2.66$\times$ & \cellcolor{famlin}2.65$\times$ & \cellcolor{famlin}2.52$\times$ & \cellcolor{famlin}2.04$\times$ & \cellcolor{famlin}1.48$\times$ & \cellcolor{famlin}1.19$\times$ & \cellcolor{famlin}\taun{4.06}{7.0} & \cellcolor{famlin}\taun{4.05}{7.0} & \cellcolor{famlin}\taun{4.07}{7.0} & \cellcolor{famlin}\taun{4.05}{7.0} & \cellcolor{famlin}\taun{4.06}{7.0} & \cellcolor{famlin}\taun{4.07}{7.0} \\
\bottomrule
\end{tabularx}

\end{table*}

% Floats placed ahead of their sections so that, at one figure or table per page, none lands among the references.
\begin{table*}[t]
\caption{\normalfont D3Tree at construction budgets $B\in\{15,31,63\}$
with the same allocation table, columns as in Table~\ref{tab:caddtree}: speedup
over AR and accepted length $\tau$ / verified draft tokens per request and
decoding round, over all 17 benchmarks. Best speedup per verifier and column
in bold.}
\label{tab:b-ablation}
\centering
% Generated by tools/gen_experiment_tables.py -- do not edit; the caption and float live in sections/experiments.tex.

\scriptsize
\definecolor{famlin}{RGB}{249,245,252}\definecolor{famtrunc}{RGB}{254,251,229}\definecolor{famtree}{RGB}{240,249,242}\definecolor{famcad}{RGB}{236,244,252}\definecolor{famours}{RGB}{254,244,233}
\ifdefined\taucw\else\newlength\taucw\newlength\tauncw\fi
\settowidth\taucw{0.00}\settowidth\tauncw{000.0}
\providecommand\taun[2]{\mbox{\makebox[\taucw][r]{#1}\ /\ \makebox[\tauncw][l]{#2}}}
\setlength{\tabcolsep}{2.6pt}
\renewcommand{\arraystretch}{0.94}
\setlength{\aboverulesep}{0.15ex}
\setlength{\belowrulesep}{0.25ex}
\begin{tabularx}{\textwidth}{ll>{\centering\arraybackslash}X>{\centering\arraybackslash}X>{\centering\arraybackslash}X>{\centering\arraybackslash}X>{\centering\arraybackslash}X>{\centering\arraybackslash}Xcccccc}
\toprule
\multirow{2}{*}[-2.7pt]{Model} & \multirow{2}{*}[-2.7pt]{Budget} & \multicolumn{6}{c}{Speedup over AR at concurrency $R$} & \multicolumn{6}{c}{Accepted length $\tau$\,/\,verified draft tokens at concurrency $R$} \\
\cmidrule(lr){3-8} \cmidrule(lr){9-14}
 & & 1 & 4 & 8 & 16 & 32 & 64 & 1 & 4 & 8 & 16 & 32 & 64 \\
\midrule
\multirow{3}{*}{Q3.5-4B} & $B{=}15$ & \textbf{3.52}$\times$ & \textbf{3.14}$\times$ & \textbf{2.66}$\times$ & \textbf{2.20}$\times$ & 1.70$\times$ & 1.37$\times$ & \taun{5.33}{15.0} & \taun{5.35}{15.0} & \taun{4.72}{7.5} & \taun{4.75}{7.5} & \taun{4.52}{5.9} & \taun{4.52}{5.9} \\
 & $B{=}31$ & 3.44$\times$ & 2.80$\times$ & 2.58$\times$ & 2.10$\times$ & \textbf{1.72}$\times$ & \textbf{1.39}$\times$ & \taun{5.77}{31.0} & \taun{5.77}{31.0} & \taun{4.72}{7.8} & \taun{4.74}{7.7} & \taun{4.52}{6.0} & \taun{4.53}{6.0} \\
 & $B{=}63$ & 3.10$\times$ & 1.81$\times$ & 2.32$\times$ & 2.00$\times$ & 1.67$\times$ & 1.35$\times$ & \taun{6.07}{63.0} & \taun{6.04}{63.0} & \taun{4.74}{8.2} & \taun{4.74}{8.1} & \taun{4.54}{6.2} & \taun{4.54}{6.2} \\
\midrule
\multirow{3}{*}{Q3.5-9B} & $B{=}15$ & \textbf{3.73}$\times$ & \textbf{3.05}$\times$ & 2.58$\times$ & \textbf{2.17}$\times$ & 1.60$\times$ & 1.28$\times$ & \taun{5.20}{15.0} & \taun{5.20}{15.0} & \taun{4.61}{7.4} & \taun{4.62}{7.4} & \taun{4.41}{5.8} & \taun{4.40}{5.9} \\
 & $B{=}31$ & 3.63$\times$ & 2.73$\times$ & \textbf{2.61}$\times$ & 2.12$\times$ & \textbf{1.63}$\times$ & \textbf{1.30}$\times$ & \taun{5.65}{31.0} & \taun{5.66}{31.0} & \taun{4.62}{7.6} & \taun{4.62}{7.6} & \taun{4.40}{6.0} & \taun{4.40}{6.0} \\
 & $B{=}63$ & 3.20$\times$ & 1.72$\times$ & 2.39$\times$ & 2.05$\times$ & 1.58$\times$ & 1.20$\times$ & \taun{5.97}{63.0} & \taun{5.99}{63.0} & \taun{4.61}{7.9} & \taun{4.64}{7.9} & \taun{4.40}{6.1} & \taun{4.42}{6.3} \\
\midrule
\multirow{3}{*}{IVL3.5-4B} & $B{=}15$ & 3.17$\times$ & \textbf{2.88}$\times$ & \textbf{2.46}$\times$ & \textbf{2.07}$\times$ & \textbf{1.66}$\times$ & 1.38$\times$ & \taun{4.79}{15.0} & \taun{4.81}{15.0} & \taun{4.80}{15.0} & \taun{4.39}{8.5} & \taun{4.19}{6.8} & \taun{4.21}{6.9} \\
 & $B{=}31$ & 3.24$\times$ & 2.69$\times$ & 1.80$\times$ & 2.02$\times$ & \textbf{1.66}$\times$ & \textbf{1.40}$\times$ & \taun{5.15}{31.0} & \taun{5.20}{31.0} & \taun{5.17}{31.0} & \taun{4.40}{9.6} & \taun{4.25}{7.6} & \taun{4.24}{7.8} \\
 & $B{=}63$ & \textbf{3.25}$\times$ & 1.82$\times$ & 1.27$\times$ & 1.89$\times$ & 1.49$\times$ & 1.28$\times$ & \taun{5.49}{63.0} & \taun{5.47}{63.0} & \taun{5.49}{63.0} & \taun{4.40}{11.9} & \taun{4.22}{9.6} & \taun{4.24}{9.6} \\
\midrule
\multirow{3}{*}{IVL3.5-8B} & $B{=}15$ & 3.66$\times$ & \textbf{3.10}$\times$ & \textbf{2.69}$\times$ & \textbf{2.25}$\times$ & \textbf{1.74}$\times$ & \textbf{1.48}$\times$ & \taun{5.08}{15.0} & \taun{5.06}{15.0} & \taun{5.06}{15.0} & \taun{4.63}{8.2} & \taun{4.45}{6.8} & \taun{4.47}{7.0} \\
 & $B{=}31$ & \textbf{3.73}$\times$ & 2.94$\times$ & 1.97$\times$ & 2.16$\times$ & 1.72$\times$ & 1.46$\times$ & \taun{5.47}{31.0} & \taun{5.48}{31.0} & \taun{5.46}{31.0} & \taun{4.62}{9.2} & \taun{4.45}{7.7} & \taun{4.49}{8.0} \\
 & $B{=}63$ & 3.54$\times$ & 2.11$\times$ & 1.53$\times$ & 2.09$\times$ & 1.60$\times$ & 1.40$\times$ & \taun{5.72}{63.0} & \taun{5.80}{63.0} & \taun{5.72}{63.0} & \taun{4.65}{11.1} & \taun{4.46}{9.3} & \taun{4.46}{9.5} \\
\midrule
\multirow{3}{*}{G4-E2B} & $B{=}15$ & 4.03$\times$ & \textbf{3.78}$\times$ & \textbf{3.16}$\times$ & \textbf{2.63}$\times$ & \textbf{2.16}$\times$ & \textbf{1.81}$\times$ & \taun{4.80}{15.0} & \taun{4.82}{15.0} & \taun{4.83}{15.0} & \taun{4.32}{8.1} & \taun{4.16}{6.5} & \taun{4.18}{6.5} \\
 & $B{=}31$ & 4.17$\times$ & 3.65$\times$ & 2.64$\times$ & 2.60$\times$ & 2.12$\times$ & 1.80$\times$ & \taun{5.17}{31.0} & \taun{5.22}{31.0} & \taun{5.19}{31.0} & \taun{4.36}{8.8} & \taun{4.18}{7.0} & \taun{4.20}{7.1} \\
 & $B{=}63$ & \textbf{4.53}$\times$ & 2.92$\times$ & 1.84$\times$ & 2.49$\times$ & 1.90$\times$ & 1.68$\times$ & \taun{5.47}{63.0} & \taun{5.47}{63.0} & \taun{5.48}{63.0} & \taun{4.34}{10.1} & \taun{4.20}{8.1} & \taun{4.19}{8.2} \\
\midrule
\multirow{3}{*}{G4-E4B} & $B{=}15$ & 3.86$\times$ & \textbf{3.36}$\times$ & \textbf{2.71}$\times$ & 2.26$\times$ & 1.76$\times$ & \textbf{1.41}$\times$ & \taun{5.12}{15.0} & \taun{5.12}{15.0} & \taun{5.12}{15.0} & \taun{4.60}{8.0} & \taun{4.42}{6.5} & \taun{4.44}{6.6} \\
 & $B{=}31$ & \textbf{4.05}$\times$ & 3.10$\times$ & 2.26$\times$ & \textbf{2.29}$\times$ & \textbf{1.77}$\times$ & \textbf{1.41}$\times$ & \taun{5.54}{31.0} & \taun{5.51}{31.0} & \taun{5.53}{31.0} & \taun{4.60}{8.7} & \taun{4.42}{7.1} & \taun{4.43}{7.3} \\
 & $B{=}63$ & 4.01$\times$ & 2.41$\times$ & 1.50$\times$ & 2.19$\times$ & 1.59$\times$ & 1.33$\times$ & \taun{5.82}{63.0} & \taun{5.83}{63.0} & \taun{5.82}{63.0} & \taun{4.62}{10.0} & \taun{4.44}{8.1} & \taun{4.44}{8.4} \\
\bottomrule
\end{tabularx}

\end{table*}

\begin{table*}[t]
\caption{\normalfont Speedup over AR and Avg.\ $\tau$ on an RTX 5090
(32\,GB) and an A800 (80\,GB), with the protocol of
Table~\ref{tab:main-speedup}. On the A800, D3Tree uses an allocation table
retuned for that GPU at $R\geq16$.}
\label{tab:device}
\centering
% Generated by tools/gen_experiment_tables.py -- do not edit; the caption and float live in sections/experiments.tex.

\footnotesize
\definecolor{famlin}{RGB}{249,245,252}\definecolor{famtrunc}{RGB}{254,251,229}\definecolor{famtree}{RGB}{240,249,242}\definecolor{famcad}{RGB}{236,244,252}\definecolor{famours}{RGB}{254,244,233}
\setlength{\tabcolsep}{2.0pt}
\renewcommand{\arraystretch}{1.12}
\begin{tabular}{llllcccccccccccccc}
\toprule
\multirow{2}{*}[-2.7pt]{GPU} & \multirow{2}{*}[-2.7pt]{Model} & \multirow{2}{*}[-2.7pt]{Strategy} & \multirow{2}{*}[-2.7pt]{Budget} & \multicolumn{7}{c}{Text benchmarks} & \multicolumn{7}{c}{Multimodal benchmarks} \\
\cmidrule(lr){5-11} \cmidrule(lr){12-18}
 &  & & & $R{=}1$ & $R{=}4$ & $R{=}8$ & $R{=}16$ & $R{=}32$ & $R{=}64$ & Avg.\ $\tau$ & $R{=}1$ & $R{=}4$ & $R{=}8$ & $R{=}16$ & $R{=}32$ & $R{=}64$ & Avg.\ $\tau$ \\
\midrule
\multirow{15}{*}{RTX 5090} & \multirow{5}{*}{Q3.5-4B} & \cellcolor{famlin}DFlash & \cellcolor{famlin} & \cellcolor{famlin}3.02$\times$ & \cellcolor{famlin}3.18$\times$ & \cellcolor{famlin}2.90$\times$ & \cellcolor{famlin}2.06$\times$ & \cellcolor{famlin}1.35$\times$ & \cellcolor{famlin}0.98$\times$ & \cellcolor{famlin}5.35 & \cellcolor{famlin}2.21$\times$ & \cellcolor{famlin}2.30$\times$ & \cellcolor{famlin}2.10$\times$ & \cellcolor{famlin}1.52$\times$ & \cellcolor{famlin}1.01$\times$ & \cellcolor{famlin}0.79$\times$ & \cellcolor{famlin}4.05 \\
 &  & \cellcolor{famtrunc}DSpark & \cellcolor{famtrunc}\multirow{-2}{*}{$N{=}15$} & \cellcolor{famtrunc}2.79$\times$ & \cellcolor{famtrunc}2.94$\times$ & \cellcolor{famtrunc}2.92$\times$ & \cellcolor{famtrunc}2.32$\times$ & \cellcolor{famtrunc}1.69$\times$ & \cellcolor{famtrunc}1.33$\times$ & \cellcolor{famtrunc}4.94 & \cellcolor{famtrunc}2.04$\times$ & \cellcolor{famtrunc}2.11$\times$ & \cellcolor{famtrunc}2.16$\times$ & \cellcolor{famtrunc}1.76$\times$ & \cellcolor{famtrunc}1.32$\times$ & \cellcolor{famtrunc}\textbf{1.11}$\times$ & \cellcolor{famtrunc}3.66 \\
\bottomrule
\end{tabular}

\end{table*}

Table~\ref{tab:caddtree} compares the strategies at every budget we ran.
No fixed budget is best across loads: the fastest fixed configuration is
DDTree ($B=15$ or $31$) at $R=1$, DDTree $B=15$ or the full DFlash draft
($N=15$) at $R=4$, and DFlash truncated to $N=7$ or $N=5$ tokens from
$R=16$ on. D3Tree with $B=15$ is faster than the best fixed
budget of each verifier at every load in 34 of the 36 verifier--load
settings; the exceptions are G4-E2B and G4-E4B at $R=1$, where DDTree
$B=31$ is ahead by at most 0.04$\times$.

CaDDTree selects the size of each tree from a single-request latency model,
so the size it selects does not depend on load. Its accepted length is
constant across loads, as is its tree size: with $B=127$ it keeps about 31
nodes per tree for Q3.5-4B and 74--81 for IVL3.5 and G4 at every load.
Larger budgets therefore raise $\tau$ by up to 0.8 tokens but lose
throughput as $R$ grows, and $B=15$ is its fastest budget at every load
of $R\geq8$. At $B=15$ the profiled latency rises so little over 15 nodes
that the cost-selected cutoff never binds: CaDDTree verifies the same full
trees as DDTree $B=15$, with the same accepted length.

D3Tree is faster than the best CaDDTree budget at every verifier and load,
by 1.23--1.47$\times$ at $R=1$, 1.34--1.48$\times$ at $R=64$, and at least
1.15$\times$ at any load. Part of the
gap at light load is implementation overhead. Our CaDDTree shares the tree
builder, verification kernels, and runtime of D3Tree, but computes its
per-request cutoff with unfused tensor operations instead of the fused
pruning kernel; at $R=1$ the median round on Q3.5-4B takes 23.0\,ms with
CaDDTree $B=15$, 17.7\,ms with DDTree $B=15$, and 15.4\,ms with D3Tree, while
all three verify the same trees. Under load, the gap reflects the selection
rule: CaDDTree keeps verifying the trees that are optimal for a single
request, whereas D3Tree spends a batch-wide budget sized to the current
load.

\subsection{Serving over Time}
\label{sec:load-trace}

The speedup tables summarize each run by one number per load. To see how
the strategies behave within a run, the scheduler records, at every
decoding step, the requests in flight and the draft tokens each of them
verifies. Fig.~\ref{fig:closed-loop} first serves 200 requests drawn from
all 17 benchmarks at a fixed concurrency on Q3.5-9B and an RTX A6000: all
are due at once, 64 are in flight, and a finished request is replaced
immediately, as in the sweeps of Table~\ref{tab:main-speedup}. D3Tree
finishes all 200 after 131.8\,s, at 1{,}301 output tokens/s, ahead of DSpark
(134.7\,s, 1{,}270 tokens/s) and AR (139.0\,s, 1{,}223 tokens/s); CaDDTree,
DFlash, and DDTree need 192--215\,s (788--884 tokens/s), because each of
their passes processes about 1{,}010 tokens against 382 for D3Tree. Its
mean latency from sending to the last token is 36.3\,s, against 36.8\,s for
DSpark, 40.0\,s for AR, and 53.0--61.3\,s for the fixed-width strategies;
AR, which advances every request by one token per pass, has the narrowest
spread and the lowest 90th percentile (50.6\,s, against 57.3\,s for D3Tree
and 57.5\,s for DSpark). The tail of the run shows the allocation at work
(Fig.~\ref{fig:closed-loop}(a)): while 64 requests are in flight D3Tree
verifies about 5 draft tokens per request, as DSpark does, but as the last
requests finish it widens again, to 7 per request below 24 requests in
flight and to whole 15-node trees for the last four, whereas DSpark keeps
its short prefixes.

A fixed concurrency holds $R$ constant, whereas in serving the load changes
within seconds. Fig.~\ref{fig:load-trace} therefore replays one arrival
trace against every strategy on the same verifier and GPU: 420 requests
arrive as a Poisson process whose rate changes over seven phases, from 0.6
to 10 requests per second, and at most 64 requests are in flight, later
arrivals waiting at the client. The trace exceeds what a single A6000
serves with this verifier under every strategy: from the burst phase on,
64 requests stay in flight and the rest queue at the client, so the
end-to-end latency is set mostly by how fast each strategy drains the
queue.

D3Tree adapts the width of every step to the load
(Fig.~\ref{fig:load-trace}(a)). With at most four requests in flight it
verifies whole 15-node trees, which this hybrid target can afford; as the
concurrency rises it narrows to 7 and then to about 5 draft tokens per
request, and it widens again whenever the batch thins out. DFlash, DDTree,
and CaDDTree verify 15 draft tokens per request throughout, so with 64
requests in flight each of their target passes processes about 1{,}010
tokens, against 383 for D3Tree and 365 for DSpark. D3Tree thus keeps the
better operating point at each end (Fig.~\ref{fig:load-trace}(b)). With at
most 8 requests in flight it accepts 4.42 tokens per round from 10.9
verified draft tokens, close to the full trees (4.52--4.57 from 15) and
0.78 more than DSpark (3.64). With 16 or more it verifies 5.2 draft tokens
per request, about as many as DSpark's 4.8, yet accepts 4.09 against 3.75,
because it keeps the most promising nodes of the whole batch rather than a
prefix of each chain.

As a result, D3Tree serves the trace with the highest output throughput,
1{,}161 tokens/s, against 1{,}145 for DSpark and 1{,}057 for AR; DFlash,
CaDDTree, and DDTree (813, 832, and 733 tokens/s) fall below AR because
their 1{,}010-token passes cost more than their acceptance returns. D3Tree
has the lowest mean latency in all seven arrival phases
(Fig.~\ref{fig:load-trace}(c)): 4--13\% below DSpark, 6--65\% below DFlash,
and 25--63\% below CaDDTree. Over the trace its mean end-to-end latency is
55.4\,s and its 90th percentile 91.1\,s, against 58.8\,s and 95.9\,s for
DSpark, 73.8\,s for AR, 118.7\,s for CaDDTree, 120.7\,s for DFlash, and
147.2\,s for DDTree. At most 167 requests are in the system with D3Tree,
against 173 with DSpark and 233--252 with the fixed-width strategies, and
its last request completes 59\,s after the last arrival, against 64\,s for
DSpark and 179--238\,s for DFlash, CaDDTree, and DDTree.

\subsection{Construction Budget}
\label{sec:ablation}

Table~\ref{tab:b-ablation} runs D3Tree with $B=31$ and $63$ under
the allocation table set for $B=15$. Up to $R_{\rm full}$ D3Tree verifies
whole trees, so a larger $B$ raises both the accepted length and the
verification width: at $R=1$, $\tau$ grows by 0.36--0.45 tokens at $B=31$
and by 0.64--0.77 at $B=63$. On IVL3.5 and G4 a single request still gains
from the larger tree, by up to 12\% with $B=31$ or $63$; on the hybrid
Qwen3.5 targets $B=15$ remains fastest. Already at $R=4$, verifying whole
larger trees costs 3--11\% of the speedup with $B=31$ and 23--44\% with
$B=63$. Above $R_{\rm full}$ the batch-wide budget $N_{\rm tot}$ caps the
verified rows, so the construction budget matters little: from $R=16$ on,
$B=31$ verifies about one more draft token per tree and stays within 5\% of
$B=15$, and $B=63$ within 12\%. A larger construction budget thus helps
only a single request on some verifiers and costs speedup at moderate load;
we keep $B=15$.

\subsection{Results on Other GPUs}
\label{sec:device-results}

Table~\ref{tab:device} repeats the comparison on an RTX 5090, for Q3.5-4B,
IVL3.5-4B, and G4-E2B, and on an A800, for all six verifiers. On the RTX 5090 the
allocation table is unchanged. On the A800, whose tiles differ from those
of the RTX 6000 Ada, we retuned the per-request allocation for $R\geq16$ on six
of the 17 benchmarks with 48 prompts each: dense targets keep $\bar n=7$ at
every $R\geq9$, and hybrid targets drop to $\bar n=4$ from $R=24$; $R\leq8$
is unchanged. Against the table of Section~\ref{sec:admission}, this
changes the speedup at $R\geq16$ by $-5\%$ to $+7\%$.

On the RTX 5090, D3Tree is the fastest strategy in 31 of the 36 settings
and 1.28--1.57$\times$ as fast as CaDDTree in every setting. On the smallest
verifier, G4-E2B, DFlash's full linear draft is ahead up to $R=8$ on the text
benchmarks (by up to 0.61$\times$, at $R=4$) and at $R=1$ on the multimodal
ones (3.09$\times$ vs.\ 2.95$\times$), and D3Tree leads from $R=16$ on; the
remaining exception is DSpark on the multimodal benchmarks of Q3.5-4B at
$R=64$ (1.11$\times$ vs.\ 1.08$\times$). At $R=64$, D3Tree reaches
1.21--1.87$\times$ over AR, whereas DDTree falls to 0.34--0.75$\times$.

On the A800, D3Tree is the fastest strategy in 61 of the 72 settings,
reaching 3.14--4.48$\times$ over AR for a single request and
1.08--1.79$\times$ at $R=64$, where DDTree falls to 0.27--0.66$\times$. It is
1.17--1.59$\times$ as fast as CaDDTree in every setting. All eleven
exceptions are DSpark: on the hybrid Qwen3.5 targets from $R=16$ on (nine
settings, by up to 0.24$\times$; Q3.5-4B, text, $R=32$: 1.67$\times$ vs.\
1.43$\times$), and on the multimodal benchmarks of IVL3.5-4B at $R=32$
(0.05$\times$) and IVL3.5-8B at $R=64$ (0.01$\times$). On Qwen3.5 the gap
does not close with the allocation: in the retuning, D3Tree stayed behind
DSpark on Q3.5-4B even when it verified as few draft tokens per request,
so it comes from the per-step cost of tree verification on this GPU, the
tree attention and the replay of the GDN states, rather than from the
number of verified tokens.

\section{Conclusion}
\label{sec:conclusion}
We presented D3Tree, which decides how much of a block-diffusion draft tree
to verify. A tree-size head with $2H{+}2$ parameters scores every candidate,
parent-chain products yield an ancestor-closed subtree, and a batch-wide node
budget aligned with the target's matrix-multiplication tiles sets how many
nodes each decoding step verifies. Under light load, D3Tree verifies whole
trees; as concurrency grows, it spends one budget across the batch instead
of a tree size fixed per request in advance. On six vision-language
verifiers and 17 text and multimodal benchmarks, D3Tree is the fastest of
the compared verification strategies in 70 of 72 load settings, reaching
3.2--4.0$\times$ over autoregressive decoding for a single request and
1.3--1.8$\times$ at 64 concurrent requests, where fixed-budget trees fall
below autoregressive speed. In serving, it completes a fixed-concurrency
workload first and attains the lowest mean latency in every phase of a
time-varying arrival trace.

Two limitations remain. On the A800, the per-step cost of tree verification
on hybrid targets, namely tree attention and the replay of recurrent
states, lets dynamically truncated chains overtake D3Tree under heavy load;
closing this gap calls for cheaper tree kernels rather than a different
allocation. In addition, our verification is greedy. Extending tree pruning
to stochastic sampling, and evaluating the optional joint training of the
drafter at scale, are left to future work.

\balance
\bibliographystyle{IEEEtran}
\bibliography{references}

\end{document}